\documentclass[final]{cvpr}

\usepackage[pdftex]{graphicx}
\usepackage{times}
\usepackage{epsfig}
\usepackage{graphicx}
\usepackage{amsmath}
\usepackage{amssymb}

\usepackage{graphbox}
\usepackage[sort, numbers]{natbib}
\usepackage{symbols}

\usepackage[ruled]{algorithm2e}
\usepackage{algorithmic}

\usepackage{color}
\usepackage{array}
\usepackage{multirow}
\usepackage{subcaption}
\usepackage{booktabs}
\usepackage{float}
\usepackage{url}
\usepackage{capt-of}
\usepackage{fixltx2e}
\usepackage{xcolor}

\usepackage{color}
\definecolor{mycitecolor}{HTML}{529C24}
\definecolor{mycitecolor1}{HTML}{668925}
\definecolor{mycitecolor2}{HTML}{E33E33}
\usepackage[pagebackref=true,breaklinks=true,colorlinks,bookmarks=false, citecolor=mycitecolor1, linkcolor=mycitecolor2]{hyperref}

\usepackage[pagebackref=true,breaklinks=true,colorlinks,bookmarks=false]{hyperref}

\makeatletter
\newcommand{\removelatexerror}{\let\@latex@error\@gobble}
\makeatother

\def\cvprPaperID{5860} %
\def\confYear{CVPR 2021}
\begin{document}

\title{SAIL-VOS 3D: A Synthetic Dataset and Baselines for Object Detection and 3D Mesh Reconstruction from Video Data}

\author{Yuan-Ting Hu, Jiahong Wang, Raymond A. Yeh, Alexander G. Schwing\\
	University of Illinois at Urbana-Champaign\\
	{\tt\small  \{ythu2, jiahong4, yeh17, aschwing\}@illinois.edu}
}

\twocolumn[{%
	\renewcommand\twocolumn[1][]{#1}%
	\maketitle
	\begin{minipage}{\textwidth}
	\vspace{-0.5cm}
	\begin{center}
		\includegraphics[width=0.245\textwidth]{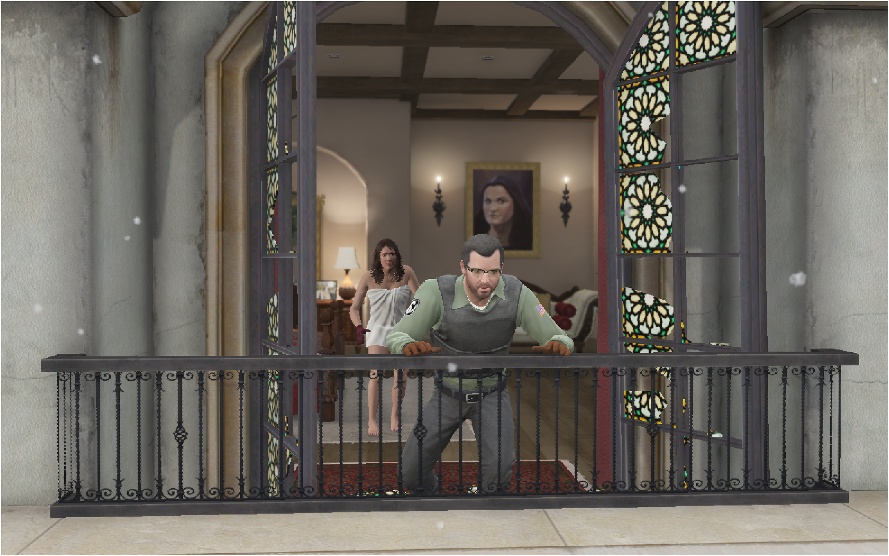} 
		\includegraphics[width=0.245\textwidth]{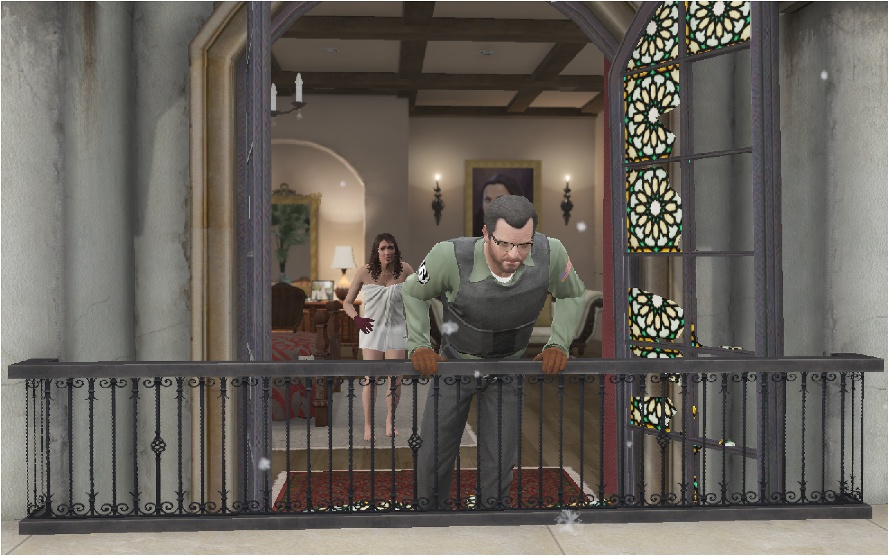} 
		\includegraphics[width=0.245\textwidth]{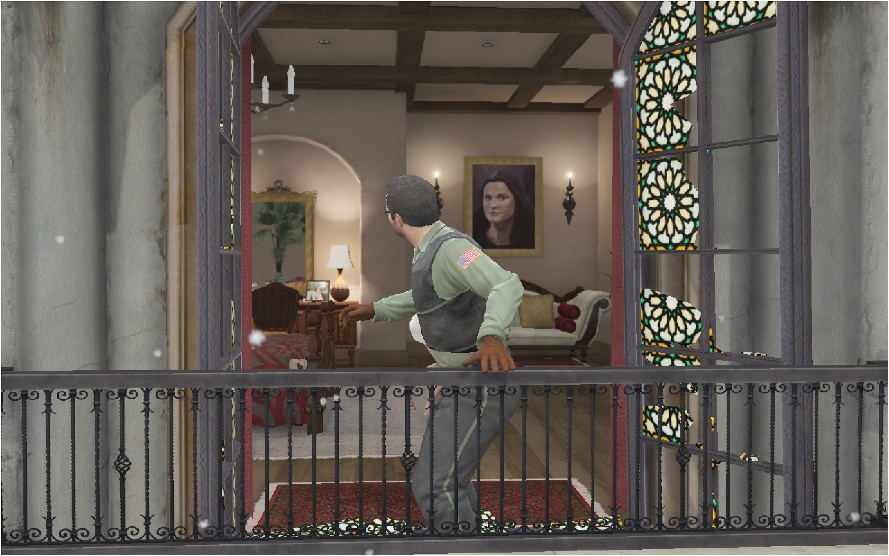} 
		\includegraphics[width=0.245\textwidth]{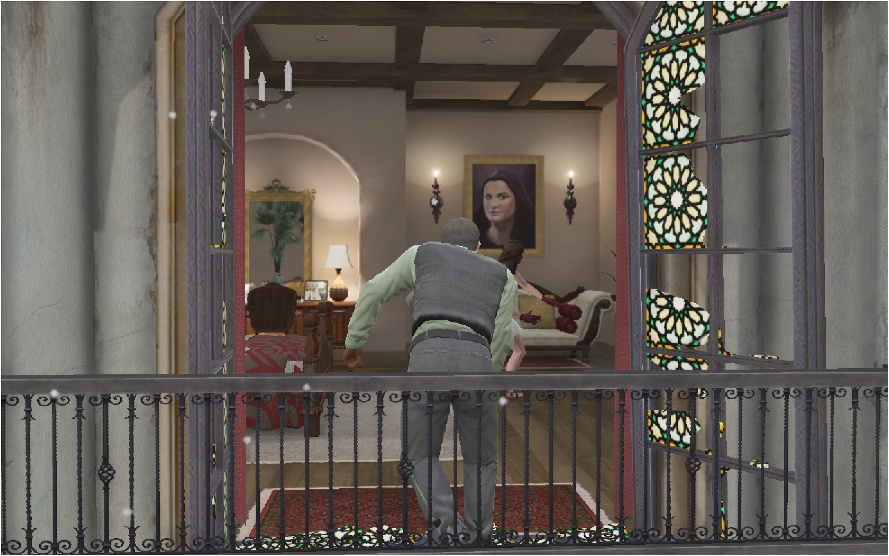} \\
		\includegraphics[width=0.245\textwidth]{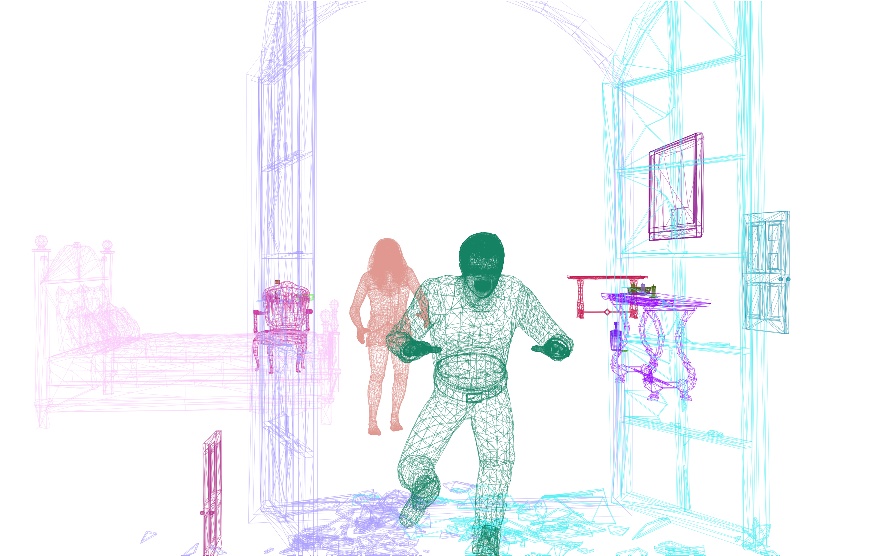} 
		\includegraphics[width=0.245\textwidth]{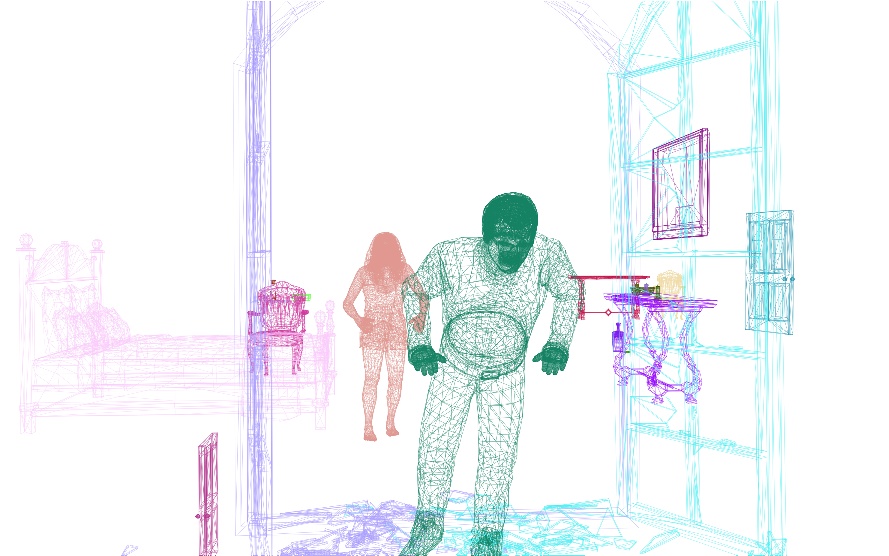} 
		\includegraphics[width=0.245\textwidth]{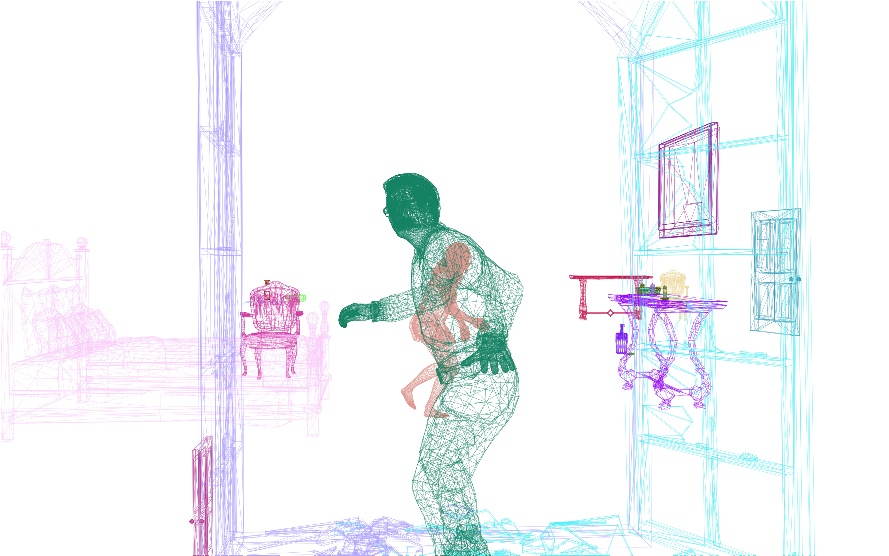} 
		\includegraphics[width=0.245\textwidth]{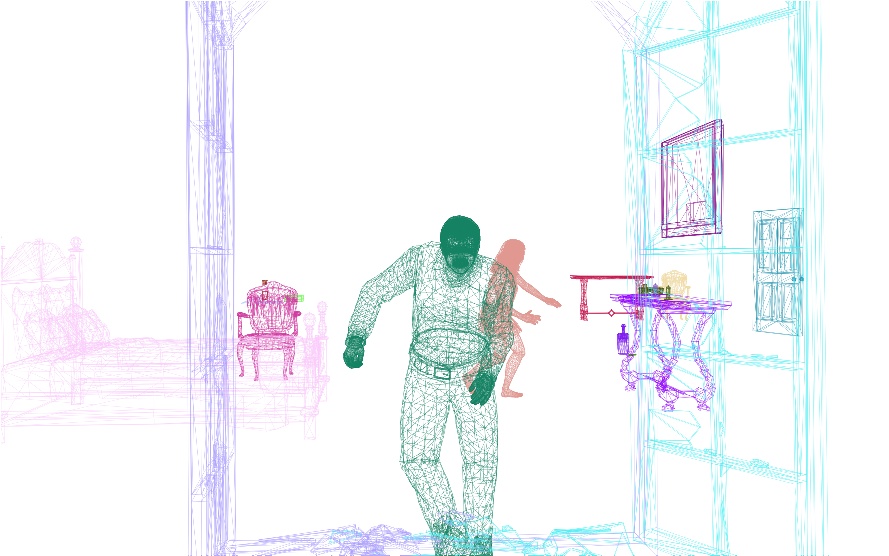}

	\end{center}
\vspace{-0.6cm}
\captionof{figure}{
Temporal information in videos provides a lot of cues for 3D mesh reconstruction. For instance, as the main character in this scene turns  we can accurately reconstruct both the front and back. Unfortunately, existing image-based datasets don't permit to develop such methods. To alleviate this we introduce the synthetic SAIL-VOS 3D dataset and develop first baselines for this task. 
}
\label{fig:teaser}
	\vspace{0.6cm}
\end{minipage}

	\vspace{-0.2cm}
}]

\begin{abstract}
Extracting detailed 3D information of objects from  video data is an important goal for holistic scene understanding. 
While recent methods have shown impressive results when reconstructing meshes of objects from a single image, results often remain ambiguous as part of the object is unobserved. 
Moreover, existing image-based datasets for mesh reconstruction don't permit to study models which integrate temporal information. To alleviate both concerns we present SAIL-VOS 3D: a synthetic video dataset with frame-by-frame mesh annotations which extends SAIL-VOS. We also develop first baselines for reconstruction of 3D meshes from video data via  temporal models. We demonstrate  efficacy of the proposed baseline on SAIL-VOS 3D and Pix3D, showing that temporal information improves reconstruction quality. Resources and additional information are available at \url{http://sailvos.web.illinois.edu}. 
\end{abstract}
\section{Introduction}

Understanding the 3D shape of an observed  object over time is an important goal in computer vision. 
Very early work towards this goal~\cite{Marr1978,Marr1980,Marr1982} focused on recovering lines and primitives like triangles, squares and circles in images. Due to many seminal contributions, the field has significantly advanced since those early days. Given a single image, exciting recent work~\cite{izadinia2017im2cad,tulsiani2018factoring,kundu20183d,gkioxari2019mesh} detects objects and infers their detailed 3D shape. Notably, the shapes are significantly more complex than the early primitives. %

To uncover the 3D shape, recently, single view 3D shape reconstruction~\cite{wang2018pixel2mesh,mahmud2020boundary,paschalidou2020learning,wu2020pq,mo2019partnet,deng2020cvxnet,chen2020bsp,peng2020convolutional} has garnered much attention. The developed data-driven and learning-based approaches achieve realistic reconstructions by inferring the 3D geometry and structure of objects. %
All those methods have in common the use of a single input image. However, moving objects and temporal information are not considered. Intuitively, as illustrated in \figref{fig:teaser}, we expect temporal information to aid 3D shape reconstruction. How can  methods benefit from complementary information available in multiple views?

\begin{table*}[t]
\vspace{-0.3cm}
\caption{Comparisons of SAIL-VOS 3D and other 3D datasets.\label{tab:dataset}} %
\vspace{-0.4cm}
\setlength{\tabcolsep}{2.5pt}
	\resizebox{\linewidth}{!}{%
		\begin{tabular}{@{} lcccccccccccc@{} }
			\toprule
			& ShapeNet~\cite{chang2015shapenet} & ModelNet~\cite{wu20153d} & PartNet~\cite{mo2019partnet} & SUNCG~\cite{song2017semantic} & IKEA~\cite{lim2013parsing} & Pix3D~\cite{sun2018pix3d} & ScanNet~\cite{dai2017scannet} & PASCAL3D+~\cite{xiang2014beyond} & ObjectNet3D~\cite{xiang2016objectnet3d} & KITTI~\cite{geiger2012we}   & OASIS~\cite{chen2020oasis}& {\bf SAIL-VOS 3D} \\
			\midrule 
			Type & Synthetic & Synthetic & Synthetic & Synthetic & Real & Real & Real & Real & Real & Real & Real & Synthetic \\
			Image/Video & - & - & - & Image & Image & Image & Video &  Image & Image & Video & Image & Video \\
			Indoor/Outdoor & - & - & - & Indoor & Indoor & Indoor & Indoor & Both & Both & Outdoor & Both & Both\\
			Dynamics & Static & Static & Static & Static & Static & Static & Static & Static & Static & Dynamic & Static & Dynamic\\
			Background & Uniform & Uniform & Uniform & Cluttered & Cluttered & Cluttered & Cluttered & Cluttered & Cluttered & Cluttered & Cluttered & Cluttered\\
			\# of Images & - & - & - & 130,269 & 759 & 10,069 & 2,492,518 & 30,362 & 90,127 & 41,778 & 140,000 & 237,611 \\
			Objects per Images & Single & Single & Single & Multiple & Single & Single & Multiple & Multiple & Multiple & Multiple & Multiple & Multiple\\	
			\# of Categories & 55 & 662 & 24 & 84 & 7 & 9 & 296 & 12 & 100 & 3 & - & 178\\
			3D Annotation Type & 3D model & 3D model & 3D model & Voxel, depth & 3D model & 3D model & Depth, 3D model & 3D model & 3D model & Point cloud & Normals, depth & 3D model, depth\\
		    Amodal/Modal GT & Amodal & Amodal & Amodal & Amodal & Amodal & Amodal & Modal & Amodal & Amodal & Modal & Modal & Amodal\\
			\# of 3D Annotations & 51,300 & 127,915 & 26,671 & 5,697,217 & 759 & 10,069 & 101,854 & 55,867 & 201,888 & 40,000 & 140,000 & 3,460,213 \\
			
			\bottomrule
		\end{tabular}
	}
	\vspace{-0.5cm}
\end{table*}

Classical multi-view 3D reconstruction~\cite{schonberger2016structure,bailey2006simultaneous1,bailey2006simultaneous2} exploits the geometric properties exposed in multiple views. For instance, structure from motion algorithms, \eg,~\cite{schonberger2016structure},  infer the 3D shape using multi-view geometry~\cite{Hartley2004}. However note that the assumption of a static scene in these classical methods is often violated in practice. Moreover, classical 3D reconstruction focuses on finding the 3D shape of observed object parts. Nonetheless it is important for methods to predict occluded parts of objects or to infer object extent beyond the observed view~\cite{palmer1999vision}. %

To achieve a more realistic and more detailed reconstruction we aim to leverage temporal information. For this we think ground-truth 3D mesh annotations for video-data are particularly helpful. Unfortunately, this form of annotated data is not available in existing datasets. 
This is not surprising as acquiring this form of data is very time-consuming and hence expensive. In addition, annotations of this form are often ambiguous due to occlusions. However, following recent work on amodal segmentation~\cite{HuCVPR2019}, we think  photo-realistic renderings of assets in computer games could come to the rescue. To study this solution, in this paper, we collect SAIL-VOS 3D, a synthetic video dataset with frame-by-frame 3D mesh annotations for objects, ground-truth depth for scenes and 2D annotations in the form of bounding boxes, instance-level semantic segmentation and amodal segmentation. We think this data serves as a good test bed for 3D perception algorithms.

In order to deal with dynamics of 3D shapes, we propose to take the temporal information into account by developing a baseline model. %
Specifically, instead of refining a spherical shape we introduce a reference mesh which carries information from  earlier frames. %

Based on the collected SAIL-VOS 3D dataset and existing image datasets like Pix3D~\cite{sun2018pix3d}, we study the proposed method as well as recent techniques for 3D reconstruction of meshes from images. 
We  illustrate the efficacy of  the developed model and the use of temporal information on the collected SAIL-VOS 3D dataset. %
Using Pix3D~\cite{sun2018pix3d} we demonstrate generality. %

\begin{figure*}[t]
\vspace{-0.7cm}
	\begin{minipage}[t]{0.79\textwidth}
		\vspace{-0.1cm}
		\begin{center}
			\includegraphics[width=\textwidth,height=2.65in]{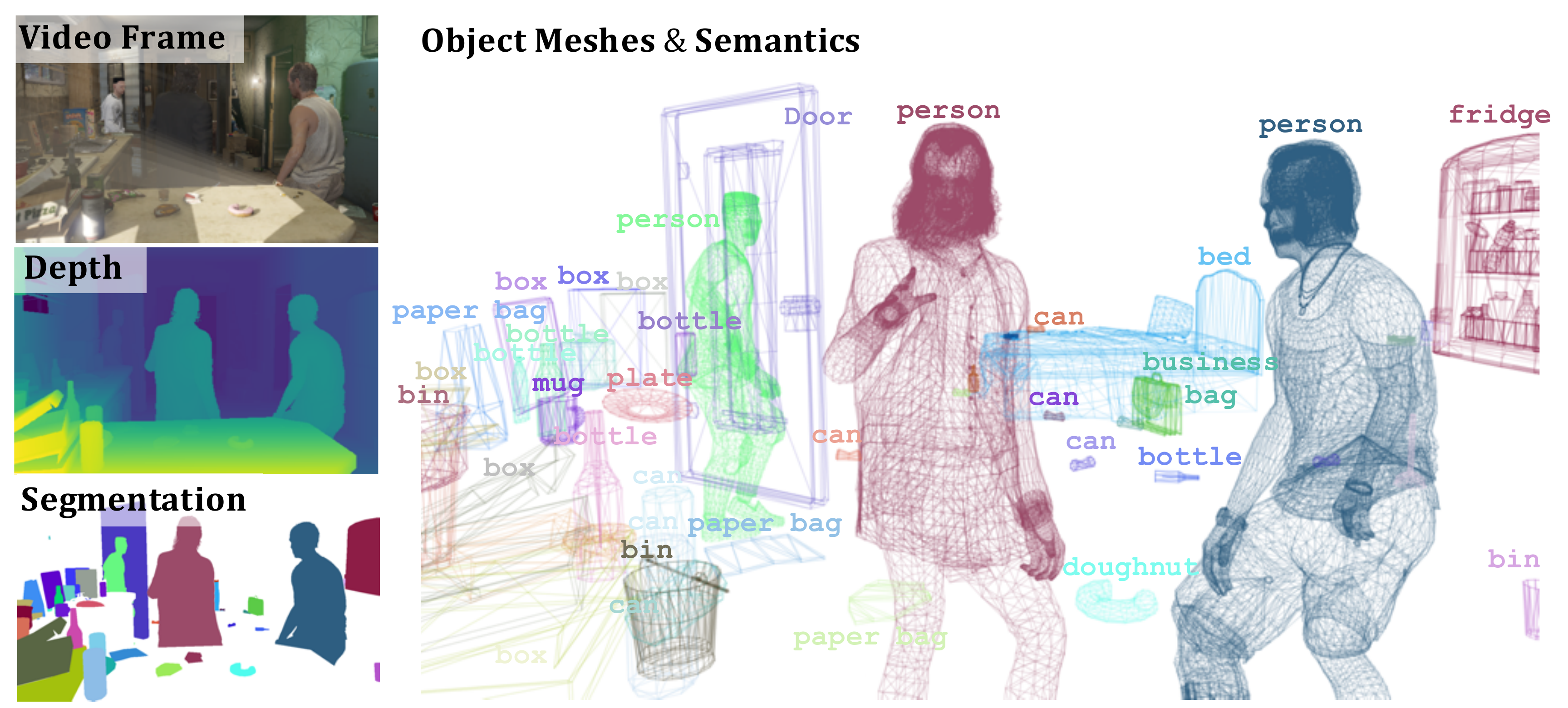}
		\end{center}
		\vspace{-0.6cm}
	\end{minipage}
	\begin{minipage}[t]{0.18\textwidth}
	\vspace{-0.1cm}
	\begin{center}
		\includegraphics[width=.9\textwidth]{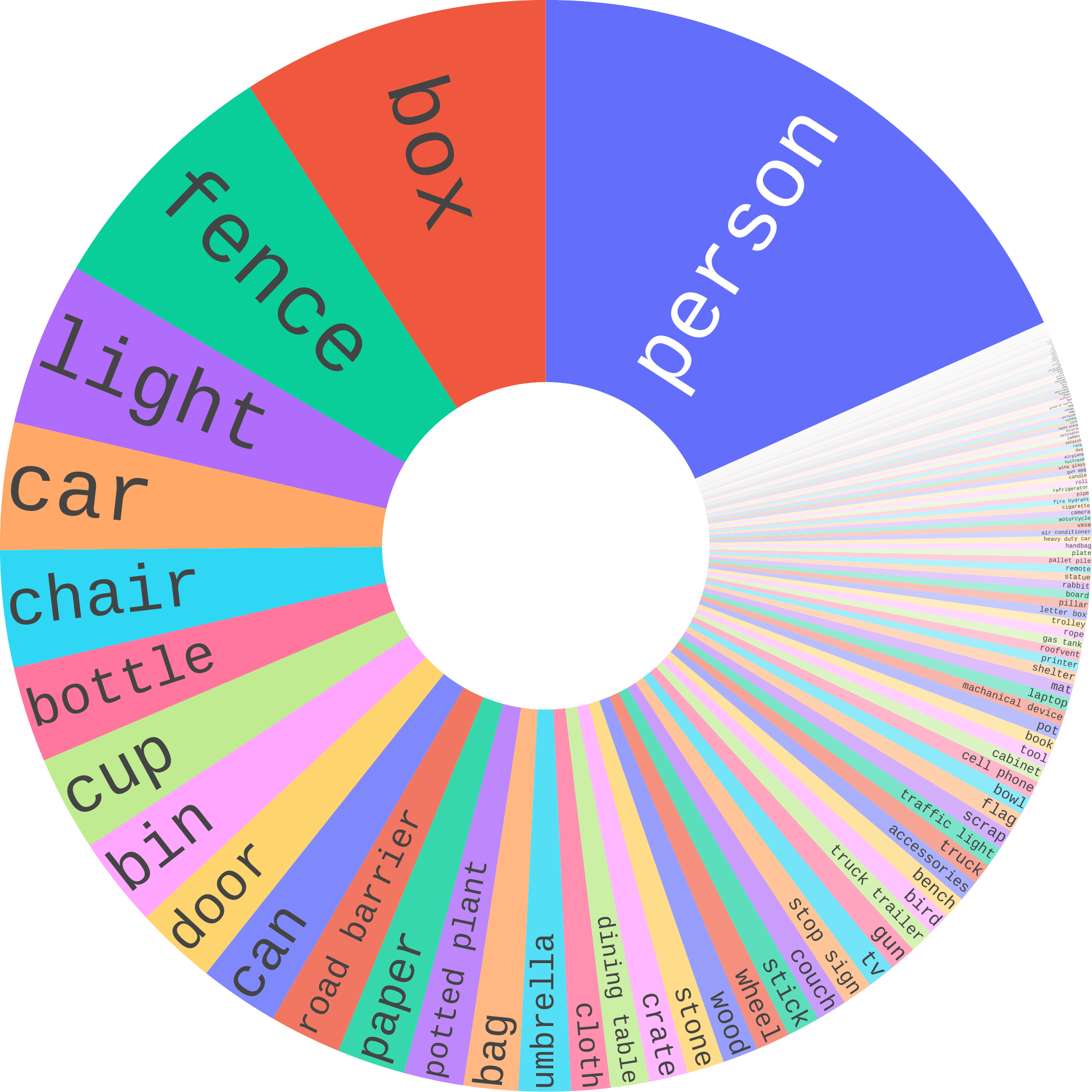}\\
		\vspace{-0.cm} Categories\\
		\includegraphics[width=.9\textwidth]{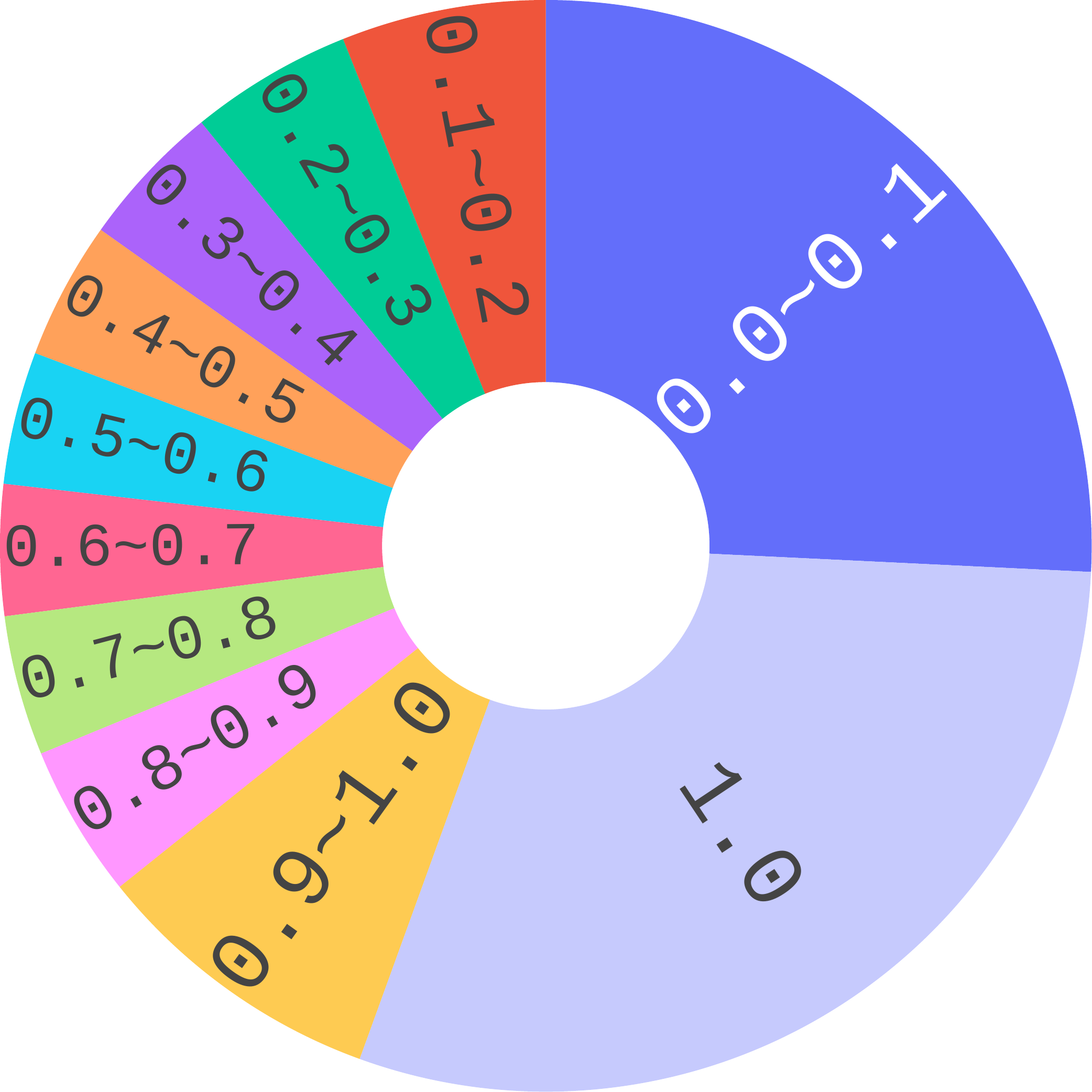}\\
		\vspace{-0.cm}Occlusion Rate\\
	\end{center}
	\vspace{-0.6cm}
\end{minipage}
\vspace{-0.1cm}
\captionof{figure}{The annotations of SAIL-VOS 3D include depth, instance-level modal and amodal segmentation, semantic labels and 3D meshes. On the right we show the statistics of the dataset in terms of numbers of objects in different categories and with different occlusion rate.}
\label{fig:stats}
\vspace{-0.4cm}
\end{figure*}

\begin{figure}[t]
	\begin{center}
		\includegraphics[width=0.19\textwidth]{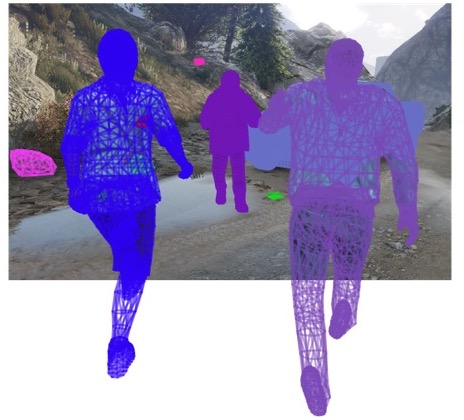}
		\includegraphics[width=0.27\textwidth]{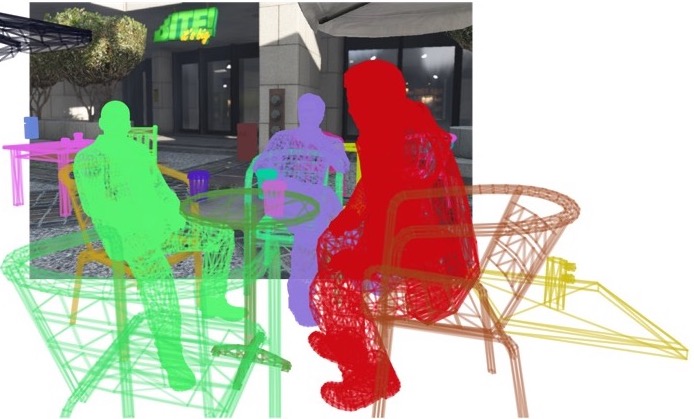}
	\end{center}
	\vspace{-0.6cm}
	\caption{We overlay the RGB frame with the 3D mesh annotations. %
	Note, the dataset contains annotations for partially observed objects including occluded and out-of-view objects. It is also a good benchmark for 2D/3D amodal perception. }
	\label{fig:outofview}
	\vspace{-0.5cm}
\end{figure}

\begin{figure*}[t]
\vspace{-0.2cm}
	\begin{minipage}[t]{\textwidth}
		\begin{center}
			\includegraphics[width=0.245\textwidth]{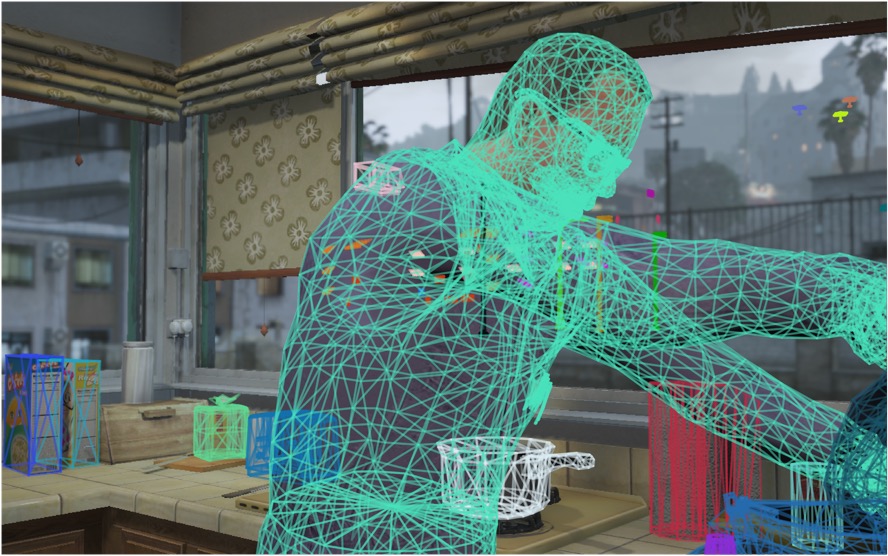}
			\includegraphics[width=0.245\textwidth]{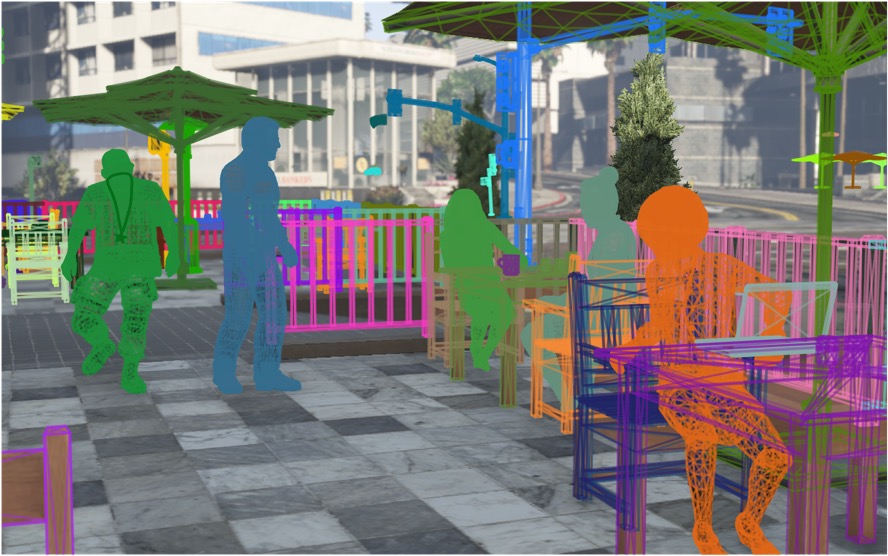}
			\includegraphics[width=0.245\textwidth]{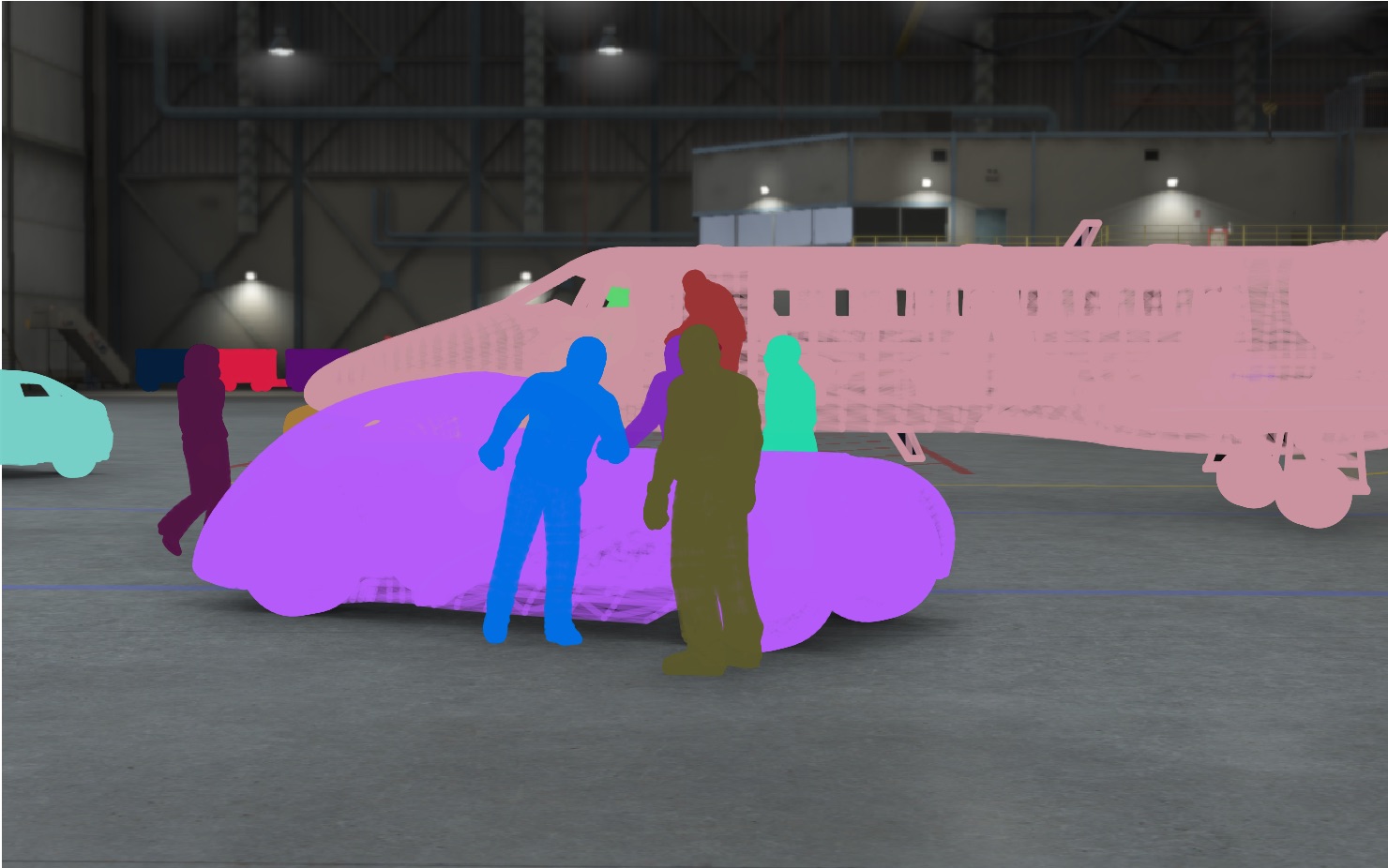}
			\includegraphics[width=0.245\textwidth]{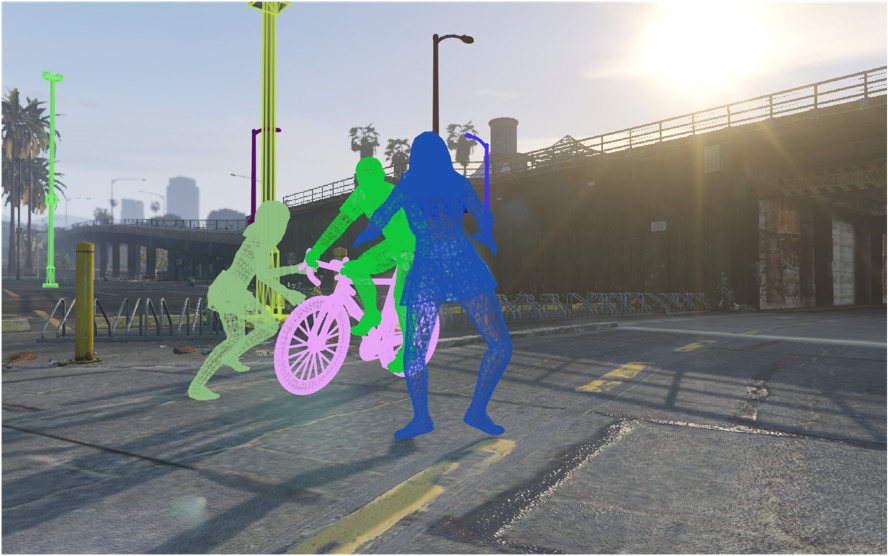}
		\end{center}
		\vspace{-0.65cm}
		\begin{center}
			\includegraphics[width=0.245\textwidth]{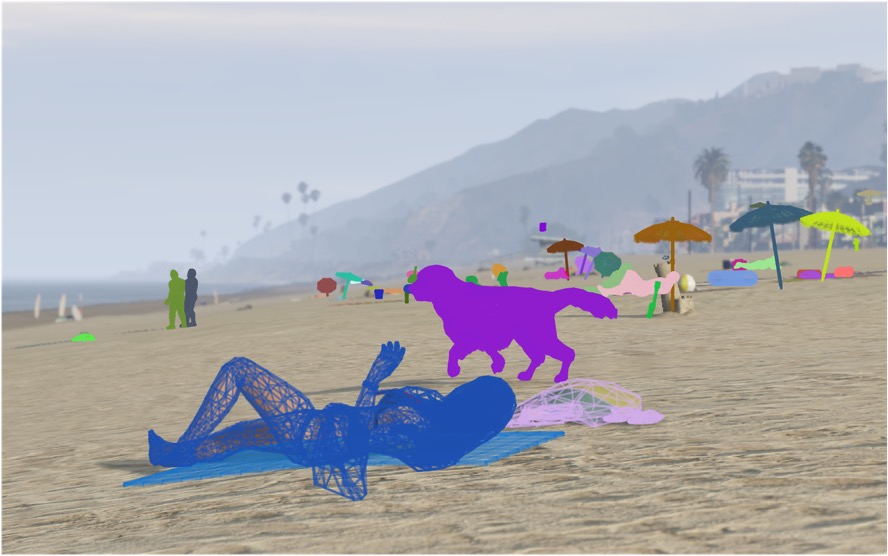}
			\includegraphics[width=0.245\textwidth]{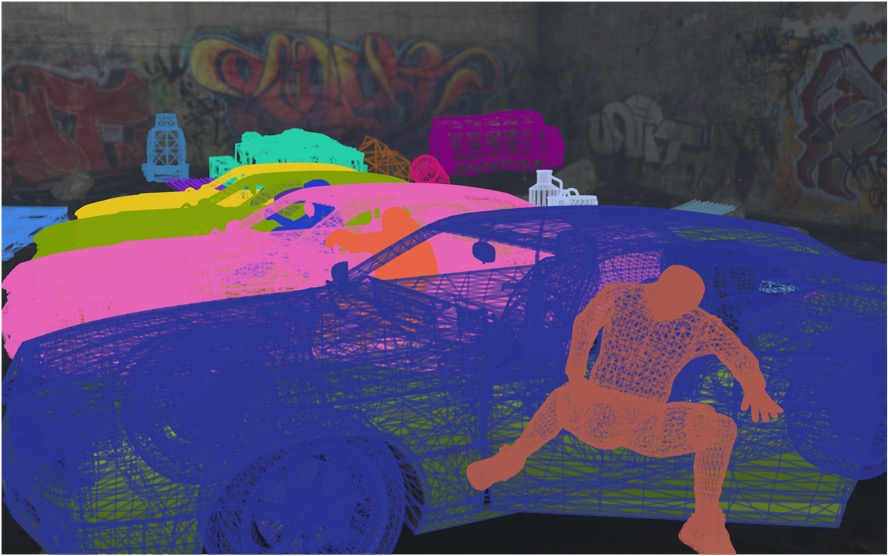}
			\includegraphics[width=0.245\textwidth]{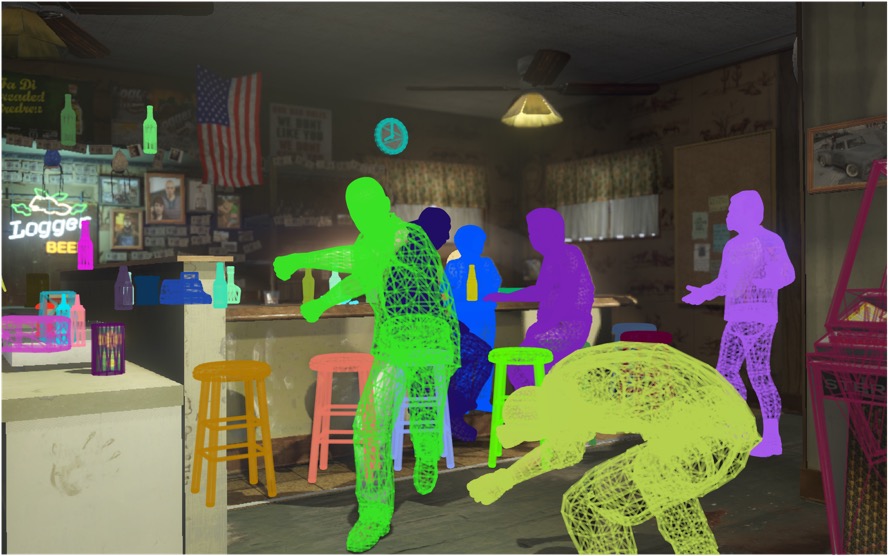}
			\includegraphics[width=0.245\textwidth]{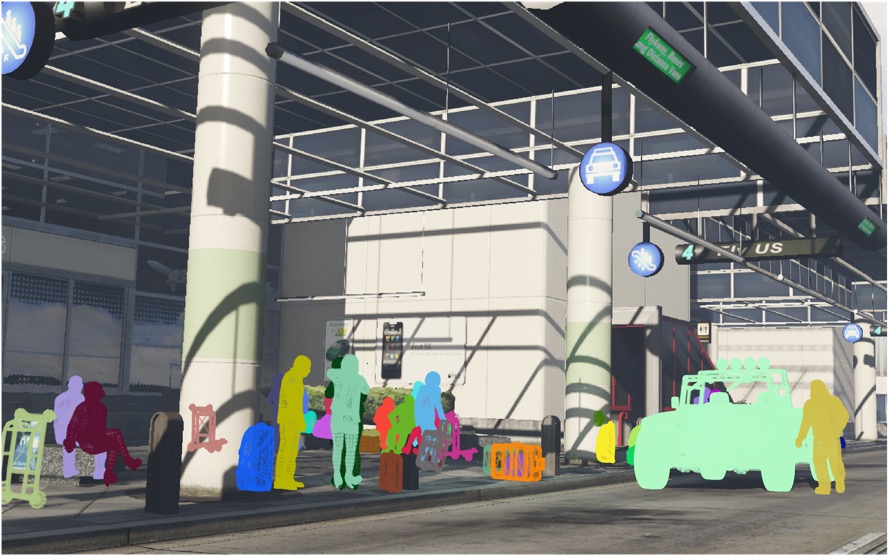}
		\end{center}
	\end{minipage}
\vspace{-0.3cm}
\captionof{figure}{Example frames overlayed with 3D mesh annotations from SAIL-VOS 3D. We use different colors for different instances.}
\label{fig:sailvos}
\vspace{-0.4cm}

\end{figure*}

\section{Related Work}\label{sec:relatedwork}

\noindent\textbf{Single-View Object Shape Reconstruction:} Recently, a number of methods~\cite{wang2018pixel2mesh,mahmud2020boundary,mescheder2019occupancy,paschalidou2020learning,wu2020pq,mo2019partnet,dai2017shape,deng2020cvxnet,chen2020bsp,peng2020convolutional,izadinia2017im2cad,zou20173d,tulsiani2018factoring,groueix2018papier,kundu20183d,dai2017shape,wu2017marrnet,nash2020polygen,mescheder2019occupancy,park2019deepsdf} have been proposed for shape reconstruction from a single image. These approaches vary by their shape representation,~\eg, meshes~\cite{wang2018pixel2mesh}, occupancy grids~\cite{wu20153d}, octrees~\cite{tatarchenko2017octree}, implicit fields~\cite{chen2019learning,mescheder2019occupancy,park2019deepsdf} and point clouds~\cite{groueix2018papier}. Many of these methods assume that there is only a single shape to be reconstructed. Consequently, these methods usually use a curated datasets like ShapeNet~\cite{chang2015shapenet}, which contains only a single object at a time. 

3D shape reconstruction methods that deal with multiple objects~\cite{izadinia2017im2cad,tulsiani2018factoring,kundu20183d,gkioxari2019mesh} usually either involve a detection network~\cite{izadinia2017im2cad,kundu20183d,gkioxari2019mesh} like Faster-RCNN~\cite{ren2015faster}, which has been very successful on 2D instance-level multi-object detection, or assume instance-level bounding boxes are given~\cite{tulsiani2018factoring}. After obtaining the bounding boxes, the 3D shape of each instance is inferred within each bounding box. To generally  deal with multiple objects, we also combine detection with shape reconstruction.  However, instead of employing a single image, the proposed approach differs in our use of video data for reconstruction.

\noindent\textbf{Multiple-View Object Shape Reconstruction:} 
3D understanding using multiple images has been studied and used for depth estimation~\cite{klodt2018supervising,teed2020deepv2d}, scene reconstruction~\cite{newcombe2011kinectfusion}, semantic segmentation for 3D scenes~\cite{kundu2020virtual}, 3D object detection~\cite{song2015joint,chen2017multi,nassar2019simultaneous}, and 3D object reconstruction~\cite{choy20163dr2n2,xie2019pix2vox}. We focus on  3D object shape reconstruction literature here as it aligns with our goal.
Traditional multi-view shape reconstruction approaches including structure from motion (SfM)~\cite{schonberger2016structure} and
simultaneous localization and mapping (SLAM)~\cite{bailey2006simultaneous1,bailey2006simultaneous2} 
 estimate 3D shapes from multiple 2D images using multi-view geometry~\cite{Hartley2004}.
However, these classical algorithms have well known constraints and are challenged by textureless objects as well as moving entities. Also,
classical multi-view geometry does not infer the shape of unseen parts of an object. %

Due to the success of deep learning, learning-based approaches to multi-view object 3D reconstruction~\cite{choy20163dr2n2,xie2019pix2vox,wiles2017silnet,tulsiani2018multi,tatarchenko2016multi,arsalan2017synthesizing,dibra2016hs,kar2017learning,huang2018deep,hu2019deep,niemeyer2019occupancy} have received a lot of attention lately. %
Again, the majority of the aforementioned methods deal with only a single object. Dealing with instance-level multi-view reconstruction remains challenging as it requires not only object detection, but also object tracking as well as 3D shape estimation. %
Closer to the goal of  our  approach, SLAM-based methods have been proposed for object-level multi-view reconstruction~\cite{civera2011towards,salas2013slam++,hu2019deep,sucar2020neural}. However, these methods are challenged by inaccurate estimation of camera poses and moving objects. In contrast, our approach aims at 3D understanding for dynamic objects. %

\noindent\textbf{3D Datasets:} A number of  3D datasets have been introduced in recent years~\cite{silberman2012indoor,geiger2012we,krause20133d,lim2013parsing,xiang2014beyond,riemenschneider2014learning,chang2015shapenet,wu20153d,xiang2016objectnet3d,armeni_cvpr16,scenenn-3dv16,armeni2017joint,song2017semantic,dai2017scannet,mccormac2017scenenet,ammirato2017dataset,sun2018pix3d,mo2019partnet,chen2020oasis}, all of which are commonly used for training, testing and evaluating 3D reconstruction algorithms. We summarize the properties of several common 3D datasets in~\tabref{tab:dataset}. Datasets annotated on real world images~\cite{lim2013parsing,sun2018pix3d,chen2020oasis,xiang2014beyond,xiang2016objectnet3d} or videos~\cite{dai2017scannet,geiger2012we} are often small  compared to synthetic datasets such as SUNCG~\cite{song2017semantic}. Moreover, KITTI~\cite{geiger2012we} and OASIS~\cite{chen2020oasis} contain only modal annotations,~\ie, only visible regions are annotated. %
In order to create large scale datasets without labor-intensive annotation processes, synthetic 3D datasets~\cite{chang2015shapenet,song2017semantic,wu20153d} are useful.

Among those, ShapeNet~\cite{chang2015shapenet} is one of the commonly used synthetic benchmarks for 3D reconstruction. However, the rendered images usually contain uniform backgrounds, making the distribution of rendered images very different from real world images.  SUNCG~\cite{song2017semantic} is one of the largest among the synthetic 3D datasets. However, the dataset contains static indoor scenes only, and doesn't cover outdoor scenarios or dynamic scenes. In fact, except for KITTI~\cite{geiger2012we}, which contains dynamic scenes, \ie, moving objects with complex motion, to the best of our knowledge, there is no large scale  3D video dataset with instance-level 3D annotation that captures dynamic scenes with diverse scenarios. 

\noindent\textbf{Synthetic Datasets:} Simulated worlds have been used for computer vision tasks such as optical flow~\cite{mayer2016large}, visual odometry~\cite{handa2014benchmark,richter2017playing} and semantic segmentation~\cite{ros2016synthia} and human shape modeling~\cite{rematas2018soccer,zhu2020reconstructing}. The game engine GTA-V has been used to collect large-scale data, \eg, for semantic segmentation~\cite{richter2016playing,richter2017playing,Kraehenbuehl_2018_CVPR}, object detection~\cite{johnson2017driving}, amodal segmentation~\cite{HuCVPR2019}, optical flow~\cite{richter2017playing,Kraehenbuehl_2018_CVPR} and human pose estimation~\cite{fabbri2018learning}.

\section{SAIL-VOS 3D Dataset}
\label{sec:data}

For more accurate 3D video object shape prediction, we propose the SAIL-VOS 3D dataset: We collect %
an object-centric 3D video dataset from the photorealistic game engine GTA-V. The game engine simulates a three-dimensional world by modeling a real-life city. 
The resulting diverse object categories and the obtainable 3D structures make the game engine a suitable choice for collecting a large-scale 3D perception dataset. 

As shown in~\figref{fig:stats} we collect the following data: 
video frames, camera matrices, depth data, instance level segmentation, instance level amodal segmentation and the corresponding 3D object shapes. Each instance is assigned a consistent id across frames.

\subsection{Dataset Statistics}
\label{sec:datastat}

The presented SAIL-VOS 3D dataset contains 484 videos with a total of 237,611 frames at a resolution of 1280$\times$800. %
Note that each video may contain shot transitions. There are a total of 6,807 clips in the dataset and on average 34.6 frames per clip. %
The dataset is annotated with 3,460,213 object instances coming from 3,576 models, which we assign to 178 categories. To assign the categories, we use the SAIL-VOS labels~\cite{HuCVPR2019}.  %
There are 48 classes that overlap with the categories in the MSCOCO dataset~\cite{lin2014microsoft}.
\figref{fig:stats} (right) shows the distribution of instances among the categories and their respective occlusion rate. Detailed statistics for clip length can be found in the Appendix. 

We compare the proposed SAIL-VOS 3D dataset with other 3D datasets in~\tabref{tab:dataset}. %
SUNCG~\cite{song2017semantic}, ScanNet~\cite{dai2017scannet}, OASIS~\cite{chen2020oasis} and SAIL-VOS 3D are large scale datasets containing more than 100K images. In OASIS only visible parts of objects are annotated. In SUNCG and ScanNet static indoor scenes are used for collecting the datasets. In contrast, SAIL-VOS 3D contains moving objects, different scenarios (both indoor and outdoor with different lighting conditions and weather), and amodal 3D annotations. Compared to the commonly used ShapeNet~\cite{chang2015shapenet}, SAIL-VOS 3D contains images rendered with cluttered backgrounds and more object categories (178 \vs 55).

We show several annotated frames sampled from SAIL-VOS 3D in \figref{fig:outofview} and \figref{fig:sailvos}. Note, the dataset contains meshes despite occlusions,~\ie,  the entire shape is annotated even if part of it is out-of-view as shown in~\figref{fig:outofview}.

\subsection{Dataset Collection Methodology}

We collect the cutscenes in GTA-V following the work of Hu~\etal~\cite{HuCVPR2019}. Similarly, we alter the weather condition, the time of day and clothing of the characters to increase the diversity of the synthetic environment. However, different from their method, we are interested in capturing 3D information in the form of meshes, camera data, \etc. This requires to extend the approach by Hu~\etal~\cite{HuCVPR2019}, and we describe the details for obtaining the data in the following. %

\noindent{\bf Meshes:} 
The shapes of objects in GTA-V are all represented as meshes. In order to retrieve the meshes, we use GameHook~\cite{Kraehenbuehl_2018_CVPR} to hook into the Direct3D 11 graphics pipeline~\cite{direct3d} which GTA-V employs. We first briefly highlight  the key components of the Direct3D 11 graphics pipeline before we delve into details.

The Direct3D 11 graphics pipeline~\cite{direct3d} consists of several stages which are traversed when rendering an image,~\eg, the vertex shader stage, where vertices are processed by applying per-vertex operations such as coordinate transformations, the stream-output stage, where vertex data can be streamed to memory buffers, the rasterizer stage, where shapes or primitives are converted into a raster image (pixels), and the pixel shader stage, where shading techniques are applied to compute the color. The intermediate output of  %
stages before the stream-output stage, \eg, the vertex shader stage, can not be directly copied to staging resources, \ie, resources which can be used to  transfer data from GPU to CPU. 
This makes access to vertex data, and hence our approach, more complicated than earlier work: design of the graphics pipeline only permits access to vertex data via the stream-output stage. %

\begin{figure*}[t]
\vspace{-0.3cm}
    \centering
    \includegraphics[width=0.95\textwidth]{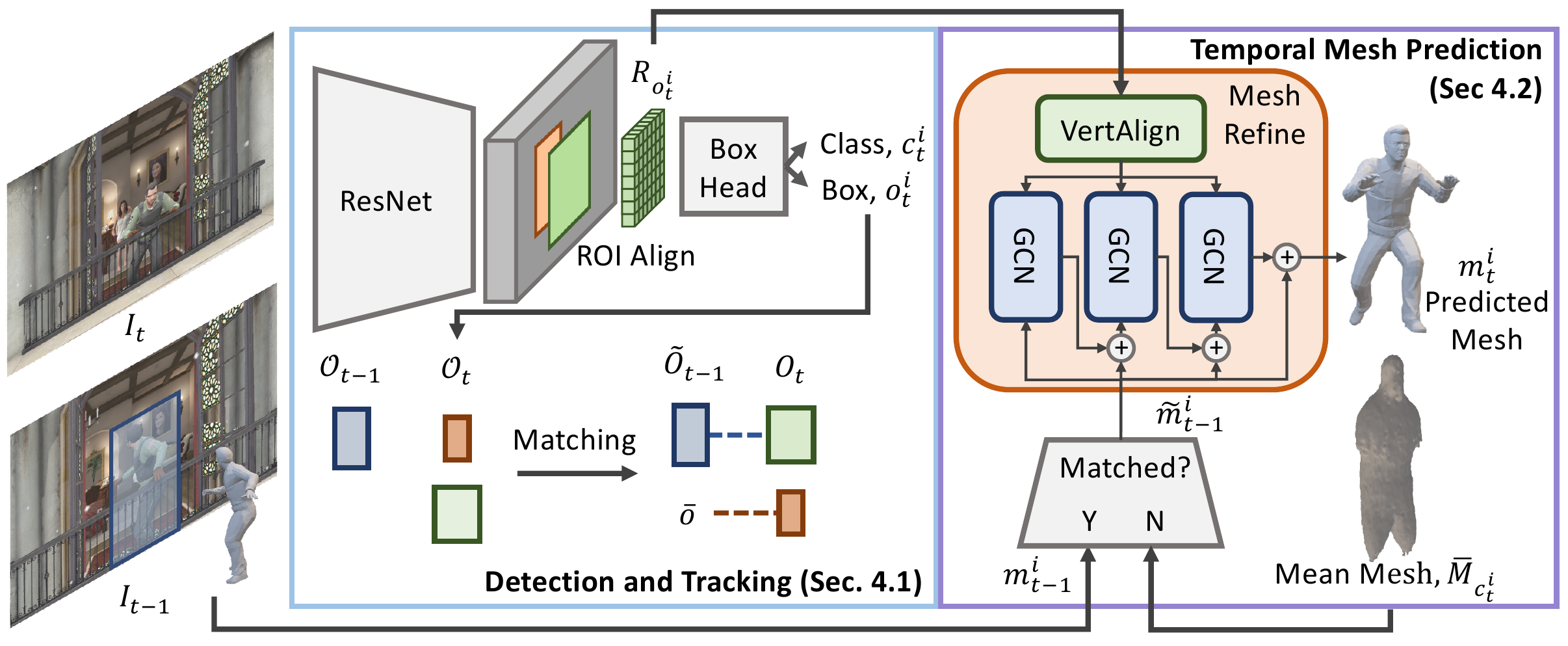}
    \vspace{-0.3cm}
    \caption{Video2Mesh: Mesh generation for the foreground person in frame $I_t$ using the boxed person in $I_{t-1}$ as a  reference.
    }
    \label{fig:overview}
    \vspace{-0.3cm}
\end{figure*}

In GTA-V, the vertex shader stage processes object vertices by applying the camera matrices as well as the transformation matrices to animate articulated models. Our goal is to retrieve the output of the vertex shader stage which contains the meshes. Since the output of the vertex shader is not directly accessible, we need to retrieve it via the stream-output stage. %
For this we enable the stream-output stage by using a hook to alter the behavior of the Direct3D function \emph{CreateVertexShader}. We ask it to not only create a vertex shader (it's original behavior) but to additionally execute %
\emph{CreateGeometryShaderWithStreamOutput}. 
This permits to append our own geometry shader 
whenever the vertex shader is created. %
Our geometry shader takes as input the output of the vertex shader. It doesn't change the vertex data and directly writes the vertex data via the stream-output stage to a stream-output buffer. As this buffer is not CPU accessible, we subsequently copy the data to a staging resource so that it can be read by the CPU. We then store the vertex data,~\ie, the meshes, %
using the  OBJ file format.

\noindent{\bf Camera \& Viewport:} 
To obtain camera matrices we collect camera data from the constant buffers in the rendering pipeline. In each draw call, we store the world matrix, which transforms model coordinates to world coordinates, the view matrix, which transforms the world coordinates to camera coordinates, the projection matrix, which transforms the camera coordinates into image coordinates, as well as the viewport, which defines the viewing region in the  rendering-device-specific coordinate.

\noindent{\bf Depth \& Others:} 
Along with the 3D shapes and camera matrices, we collect depth by copying the depth buffer from the rendering pipeline. We label shot transitions for each video. For each object in the dataset, we also collect additional information like the 2D segmentation, 2D amodal segmentation and 2D/3D human poses following Hu~\etal~\cite{HuCVPR2019}.  
\section{Video2Mesh for 3D reconstruction}

Given a video sequence $(I_1, \hdots, I_T)$  of $T$ frames, we are interested in inferring the set of meshes $\cM_t$ for the existing objects in each frame $I_t$.
A pictorial overview of our approach, Video2Mesh, is shown   in~\figref{fig:overview}. 

To infer the meshes, we first detect objects in each frame $I_t$.  We let the unordered set $\cO_t = \{o_t^i\}$ refer to the detected objects, \ie, $o_t^i$ is the $i^{\text{th}}$ detected object. Note that the set is unordered, \ie, $o_t^i$ and $o_{t-1}^i$ may not refer to the same object in frames $I_t$ and $I_{t-1}$. 
To establish  object correspondence across time, we solve an assignment problem. This permits to use  temporal information during shape prediction. Formally, for every frame $I_t$ we rearrange the   objects $\cO_{t-1}$  from the previous frame into the  ordered tuple $\tilde{\cO}_{t-1}^t = (\tilde{o}_{t-1}^1, \hdots, \tilde{o}_{t-1}^{n_t})$, where $n_t = |\cO_t|$. 
Specifically, for object $o_t^{i}$ from frame $I_t$ we let $\tilde{o}_{t-1}^i$ denote the corresponding object from frame $I_{t-1}$. %
If we cannot find a match in the previous frame, a special token is assigned to $\tilde{o}_{t-1}^i$. 

With the objects tracked, we can use the prediction from the previous frame as a reference to predict meshes in the current frame. %
We detail  detection and tracking in~\secref{sec:detrack} and explain the temporal mesh prediction in~\secref{sec:cond}.

\subsection{Detection and Tracking}
\label{sec:detrack}
To obtain the set of objects $\cO_t$ for frame $I_t$, we first use Faster-RCNN~\cite{ren2015faster} to detect  object bounding boxes and their class. Hence, each object $o_t^i = (b_t^i, c_t^i)$ consists of a bounding box and class estimate, \ie, $b_t^i$ and $c_t^i$. As previously mentioned, there are no correspondences for the detected objects between frames. To obtain those we solve an assignment problem. %

Formally, given the detected objects from two consecutive frames, the assignment problem is formulated as follows: 
\bea\nonumber
\max_{\mA_t}& \sum_{i,j} \mA_{t, ij} \left(\text{IoU}(b_t^i, b_{t-1}^j) - L(c_t^i, c_{t-1}^j)\right)\\ \;\;\text{s.t.}\;& \mA_{t,ij} \in \{0,1\}, \sum_{i}\mA_{t, ij}\leq1, \sum_{j}\mA_{t, ij}=1.
\eea
Here, $\mA_t \in \{0,1\}^{n_t \times n_{t-1}}$ %
is an assignment matrix, IoU is the intersection over-union distance between two bounding boxes and $L$ is the zero-one loss between two class labels. Intuitively, detected objects from two different frames will be matched if their bounding boxes are spatially close and their class labels are the same. %

Given the assignment matrix $\mA_t$, we  align the objects of two frames as follows: %
\bea
\begin{array}{ll}
\tilde{o}_{t-1}^i \leftarrow o_{t-1}^j, &\text{if}\;\; \mA_{t, ij} = 1 \;\land \; \text{IoU}(b_t^i, b_{t-1}^j) > 0.5,\\
\tilde{o}_{t-1}^i \leftarrow \bar{o}, & \text{otherwise}.
\end{array}
\label{eq:trackassign}
\eea
Here, $\bar{o}$ denotes a special token indicating that there were no matches. We don't match objects if their  IoU is too low.  %

\subsection{Temporal Mesh Prediction}%
\label{sec:cond}
Our goal is to predict the set of meshes $\cM=\{m_t^i\}$ for all objects $o_t^i \in \cO_t$. In our case, each mesh $m_t^i$ is a \textit{triangular mesh}, \ie, $m_t^i=(V_t^i,F_t^i)$ is characterized by a set of $K$ vertices $V_t^i \in \mathbb{R}^{K \times 3}$ and a set of triangle-faces $F_t^i \subseteq [1, \dots, K]^3$. %

To predict a mesh $m_t^i$, we deform the vertices of a reference mesh $\tilde{m}_{t-1}^i$ based on the detected object's ROI feature $R_{o_t^i}$ extracted from the image frame. %
The reference mesh $\tilde{m}_{t-1}^i$ is  a class-specific mean mesh $\bar{M}_{c_t^i}$ if object $o_t$ is not tracked, and it is the predicted mesh $m_{t-1}^i$ from the previous frame if object $o_t$ is tracked. Formally, 
\bea\label{eq:reference}
\tilde{m}_{t-1}^i =
\begin{cases}
\mT_{o_t^i} (\bar{M}_{c_t^i}) & \text{if } \tilde{o}_{t-1}^i = \bar{o},\\
       m_{t-1}^i & \text{otherwise}.
\end{cases}  
\eea
Recall, $\bar{o}$ is a special token indicating that there are no matches from the previous frame (see \equref{eq:trackassign}).
We use the transformation matrix $\mT_{o_t^i}$ to align the mean mesh $\bar{M}_{c_t^i}$ with the reference frame of $o_t^i$. We parameterize $\mT_{o_t^i}$ using a deep-net, which regresses the transformation matrix from the ROI features of $o_t^i$. Specifically, a mapping from $R_{o_t^i}$ to a rotation matrix with three degrees of freedom (the three Euler angles) is inferred. We do not consider translation here, as the mesh is centered in a bounding box.

Next, following prior work~\cite{wang2018pixel2mesh, gkioxari2019mesh} we refine the vertices of the reference mesh $\tilde{m}_{t-1}^i$ by performing $L$ stages of refinement using the {\tt MeshRefine} procedure summarized in Module~\ref{mo:refine}. Different from prior work, our reference mesh is propagated from the previous frames, \ie, 
\bea
\tilde{V}_t^{(l+1),i} = {\tt MeshRefine}^{(l)}(R_{o_t^i}, \tilde{V}_t^{(l),i}),
\eea
where $\tilde{V}_t^{(0),i}$ are the vertices of the reference mesh $\tilde{m}_{t-1}^i$.
Finally, we use the original reference's faces $\tilde{F}_{t-1}^i$ for the final mesh output, \ie,
\be
m_t^i = (V_t^{(L),i}, \tilde{F}_{t-1}^i).
\ee
In greater detail, the {\tt MeshRefine} procedure in Module~\ref{mo:refine} extracts image-aligned feature vectors for each vertex from an object's ROI feature $R_{o_t^i}$ using the {\tt VertAlign} operation~\cite{wang2018pixel2mesh,gkioxari2019mesh}, where $E^{(l)}$ denotes the number of feature dimensions. 
Subsequently, a Graph Convolution Network (GCN) is used to learn a refinement offset $\Delta_V$ to update the reference vertices $\tilde{V}$. Finally, the module returns the updated mesh vertices to be processed in the next stage. In our case, we use $L=3$ and each $\text{GCN}^{(l)}$ consists of three GraphConv layers with ReLU non-linearity. 

\begin{table}[t]
\begin{minipage}[t]{\linewidth}
\begingroup
\removelatexerror%
\renewcommand{\algorithmcfname}{Module}
\begin{algorithm*}[H]
\caption{$\tt{MeshRefine}^{(l)}(R, \tilde{V})$}
\vspace{-0.05cm}
\label{mo:refine}
\textbf{Input:} ROI Feature $R$ and reference vertices $\tilde{V}$\\
$\phi_V = {\tt VertAlign}({\tt Conv1x1}^{(l)}(R)) \in \mathbb{R}^{K \times E^{(l)}}$\\
$\Delta_V = {\tt GCN}^{(l)}(\text{Concat}(\phi_V, \tilde{V})) \in \mathbb{R}^{K\times 3}$\\
\textbf{Return:} $\tilde{V} + \Delta_V$
\end{algorithm*}
\endgroup
\end{minipage}
\vspace{-0.5cm}
\end{table}

\begin{table*}[t]
	\vspace*{-0.2cm}
	\caption{Performance on SAIL-VOS 3D. Following MeshR-CNN~\cite{gkioxari2019mesh} we report box AP, mask AP, and mesh AP. %
		\label{tab:exp}} %
	\vspace*{-0.15in}
	\resizebox{\linewidth}{!}{%
		\begin{tabular}{@{\extracolsep{3pt}} lcccccccccccccccccc @{} }
			\toprule
			\multirow{2}{*}{\%} & \multicolumn{3}{c}{All} & \multicolumn{3}{c}{Small Objects} & \multicolumn{3}{c}{Medium Objects} & \multicolumn{3}{c}{Large Objects}\\ %
			\cline{2-4} \cline{5-7} \cline{8-10} \cline{11-13} \cline{14-16} \cline{17-19}
			& $\text{AP}^{\text{box}}$ & $\text{AP}^{\text{mask}}$ & $\text{AP}^{\text{mesh}}$ & $\text{AP}^{\text{box}}$ & $\text{AP}^{\text{mask}}$ & $\text{AP}^{\text{mesh}}$ & $\text{AP}^{\text{box}}$ & $\text{AP}^{\text{mask}}$ & $\text{AP}^{\text{mesh}}$ & $\text{AP}^{\text{box}}$ & $\text{AP}^{\text{mask}}$ & $\text{AP}^{\text{mesh}}$\\
			\midrule

			Pixel2Mesh$\dagger$ & \multirow{4}{*}{25.29}          & \multirow{4}{*}{23.08}          & 8.96           & \multirow{4}{*}{6.47} & \multirow{4}{*}{3.81} & 2.30          & \multirow{4}{*}{21.72}          & \multirow{4}{*}{19.64}& \bf 10.73         & \multirow{4}{*}{44.35} & \multirow{4}{*}{43.78}          & 19.09         \\
			Pix2Vox$\dagger$ &        &         & 6.62           &  &  & 2.81          &           &  & 7.51         &  &           & 16.56          \\
			MeshR-CNN~\cite{gkioxari2019mesh} &        &          & 9.68           & &  & 2.53          &           &  & 10.35          &  &           & 20.80         \\
			Video2Mesh (Ours) &        &          & \bf 10.21          & &  & \bf 2.93          &           &  & 9.78          &  &    & \bf 22.56          \\
			\bottomrule
		\end{tabular}
	}
	\resizebox{\linewidth}{!}{%
		\begin{tabular}{@{\extracolsep{3pt}} lcccccccccccccccccc @{} }
			\toprule
			\multirow{2}{*}{\%} & \multicolumn{3}{c}{Slightly Occluded} & \multicolumn{3}{c}{Heavily Occluded} & \multicolumn{3}{c}{Short Clips} & \multicolumn{3}{c}{Long Clips}\\ %
			\cline{2-4} \cline{5-7} \cline{8-10} \cline{11-13} \cline{14-16} \cline{17-19}
			& $\text{AP}^{\text{box}}$ & $\text{AP}^{\text{mask}}$ & $\text{AP}^{\text{mesh}}$ & $\text{AP}^{\text{box}}$ & $\text{AP}^{\text{mask}}$ & $\text{AP}^{\text{mesh}}$ & $\text{AP}^{\text{box}}$ & $\text{AP}^{\text{mask}}$ & $\text{AP}^{\text{mesh}}$ & $\text{AP}^{\text{box}}$ & $\text{AP}^{\text{mask}}$ & $\text{AP}^{\text{mesh}}$\\
			\midrule 
			Pixel2Mesh$\dagger$ & \multirow{4}{*}{30.10}          &  \multirow{4}{*}{29.90}          & 12.44        & \multirow{4}{*}{16.27} & \multirow{4}{*}{13.04 }        & 2.21          & \multirow{4}{*}{29.38} & \multirow{4}{*}{26.85} & 12.92         & \multirow{4}{*}{23.33} & \multirow{4}{*}{22.74}         & 8.43          \\
			Pix2Vox$\dagger$ &           &            & 9.31          &  &          & 2.80          &  &  & 9.05          &  &          & 6.80           \\
			MeshR-CNN~\cite{gkioxari2019mesh} &           &            & 12.66          &  &          & 3.34          &  &  & \bf 13.39          &  &          & 9.03           \\
			Video2Mesh (Ours) &           &            & \bf 13.58          &  &          & \bf 3.62        &  &  & 12.85         &  &          & \bf 9.26           \\

			\bottomrule
		\end{tabular}
	}
	\vspace{-0.1cm}
\end{table*}

\begin{table}[t]
	\centering
	\caption{Ablation study of our method. ``MS'': mean shape. ``T'': estimate $\mT$ (\secref{sec:cond}). ``Temp.'': temporal prediction (\secref{sec:cond}).}
	\label{tab:ablation}
	\vspace*{-0.15in}
	\resizebox{0.8\linewidth}{!}{%
		\begin{tabular}{@{} ccccccccccccc@{} }
			\toprule
			&MS & T & Temp. & $\text{AP}^{\text{box}}$ & $\text{AP}^{\text{mask}}$ & $\text{AP}^{\text{mesh}}$  \\
			\midrule 
			{\footnotesize 1)}&-&-&-& 25.29 & 23.08 & 8.96\\
			{\footnotesize 2)}&\checkmark&-&-& 25.29 & 23.08 & 9.23\\
			{\footnotesize 3)}&\checkmark&\checkmark&-& 25.29 & 23.08 & 9.66\\
			{\footnotesize 4)}&\checkmark&\checkmark&\checkmark& 25.29 & 23.08 & {\bf 10.21}\\
			
			\bottomrule
		\end{tabular}
	}
	\vspace{-0.4cm}
\end{table}

\begin{figure*}
\vspace{-0.2cm}
	\begin{minipage}[t]{\textwidth}
	\setlength{\tabcolsep}{1pt}
    \renewcommand{\arraystretch}{0.95}
        \begin{tabular}{ccccc}
        ~\cite{gkioxari2019mesh} &
        \includegraphics[align=c, width=0.235\textwidth]{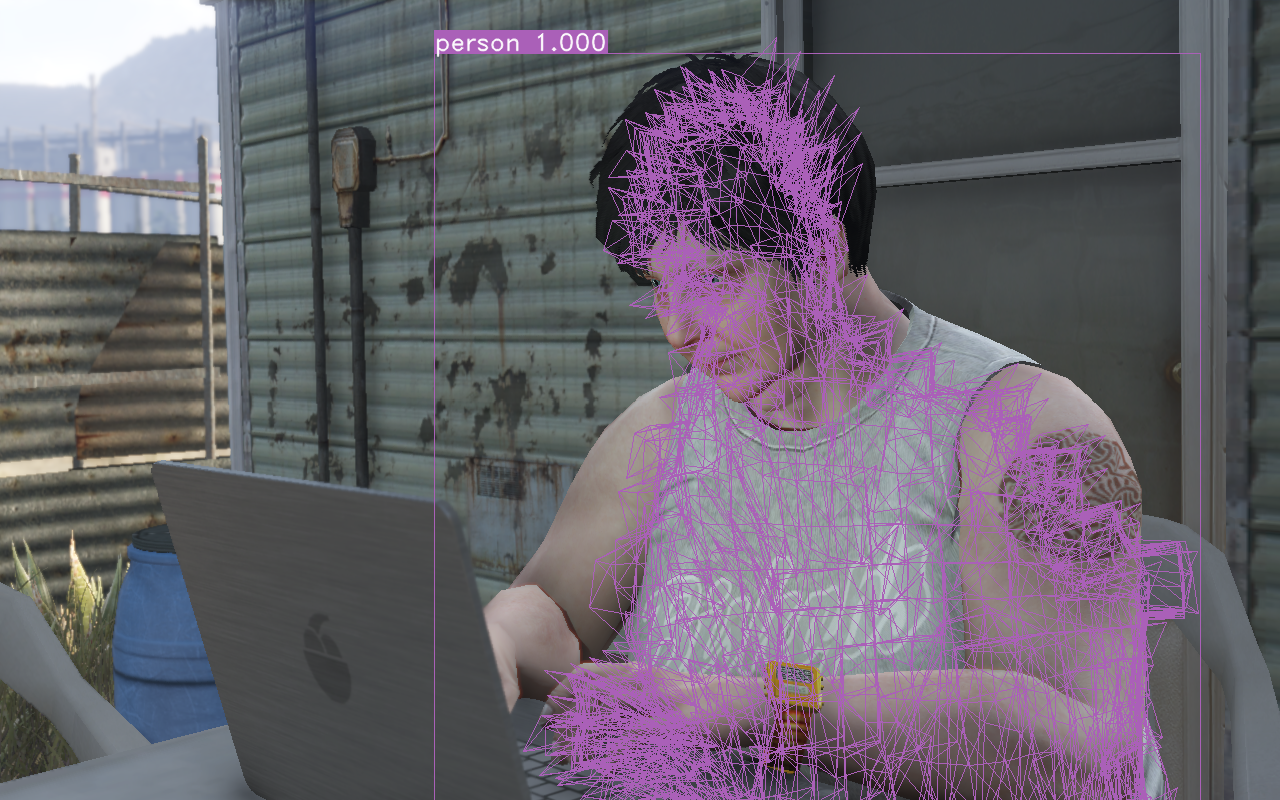} & 			\includegraphics[align=c, width=0.235\textwidth]{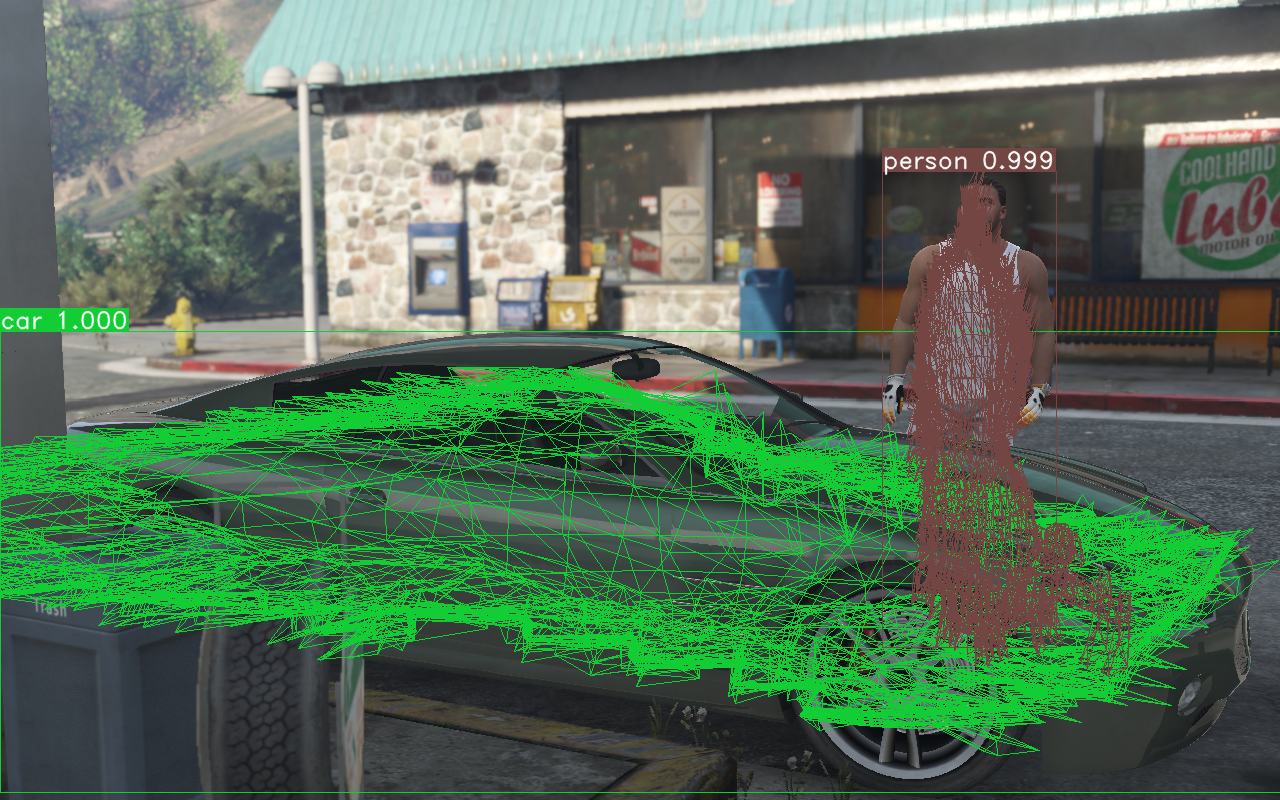} & 			\includegraphics[align=c, width=0.235\textwidth]{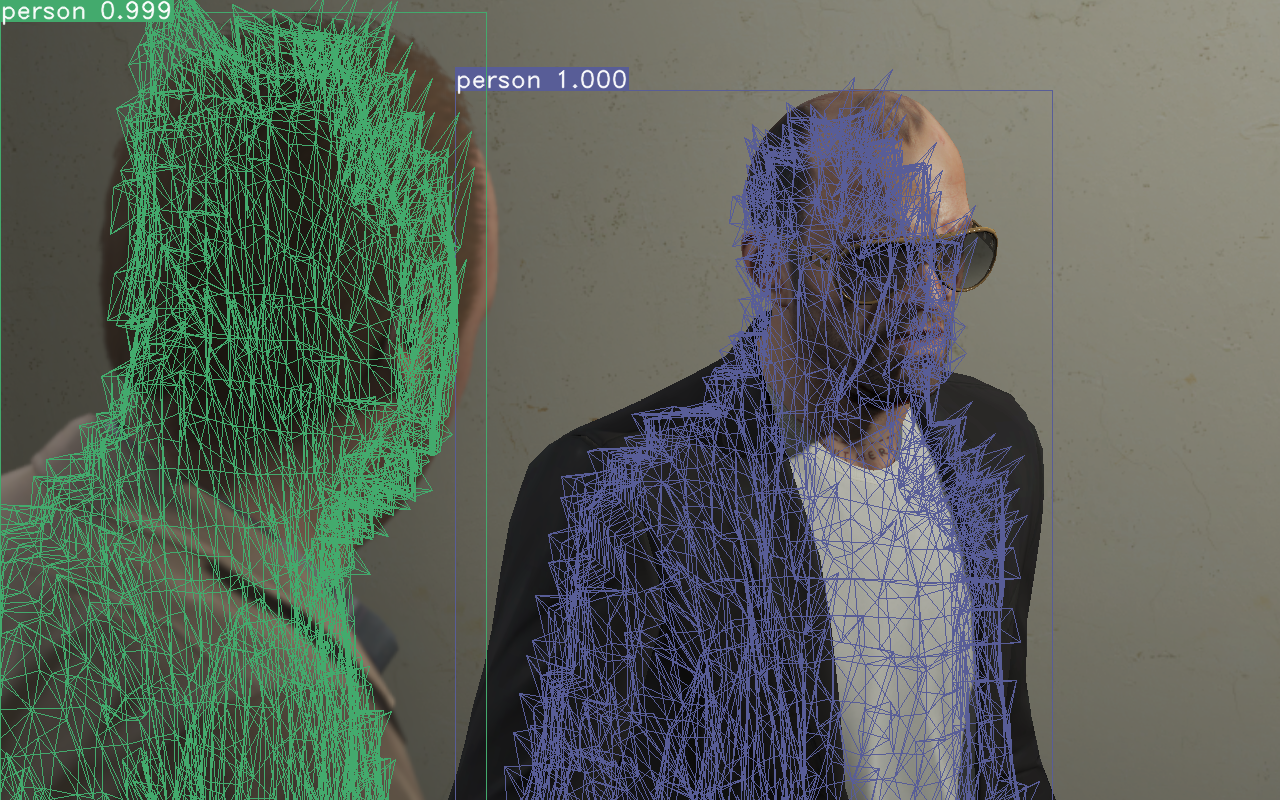} & 			\includegraphics[align=c, width=0.235\textwidth]{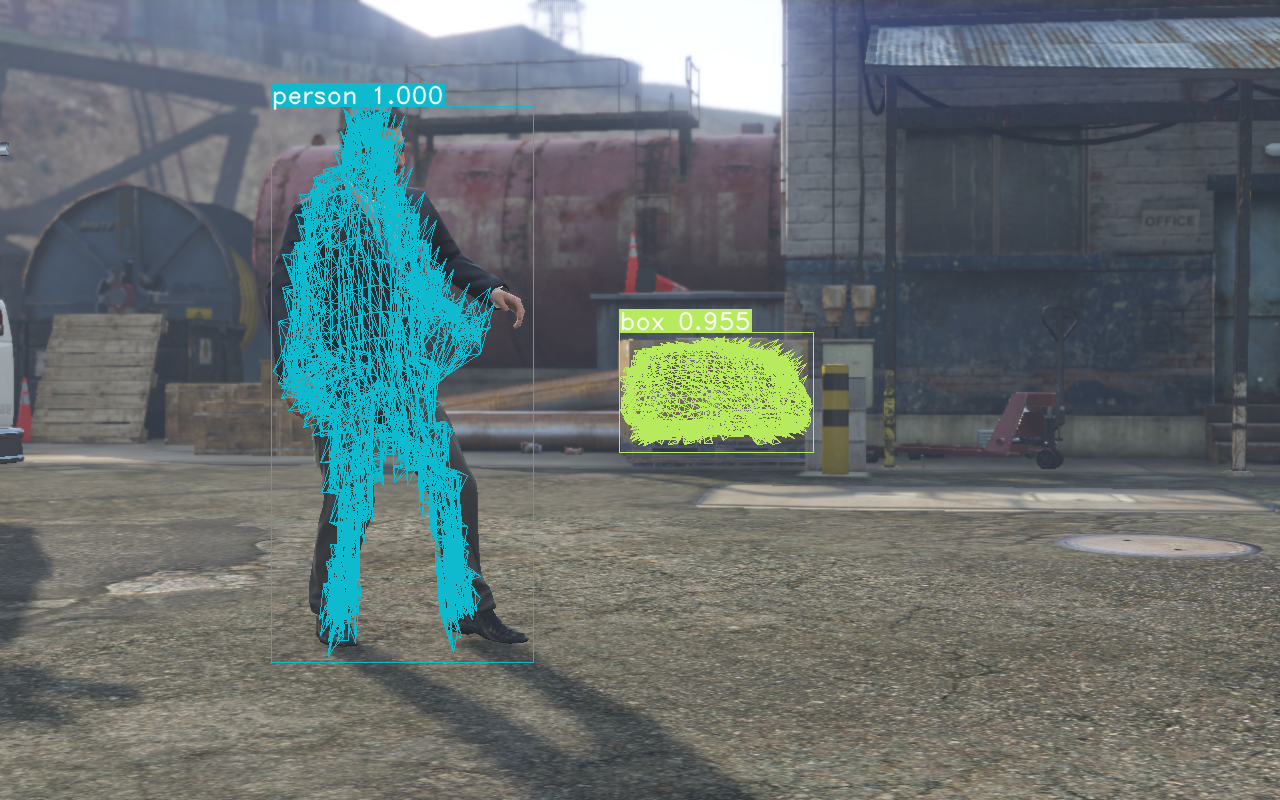} \\
        Ours & 			\includegraphics[align=c, width=0.235\textwidth]{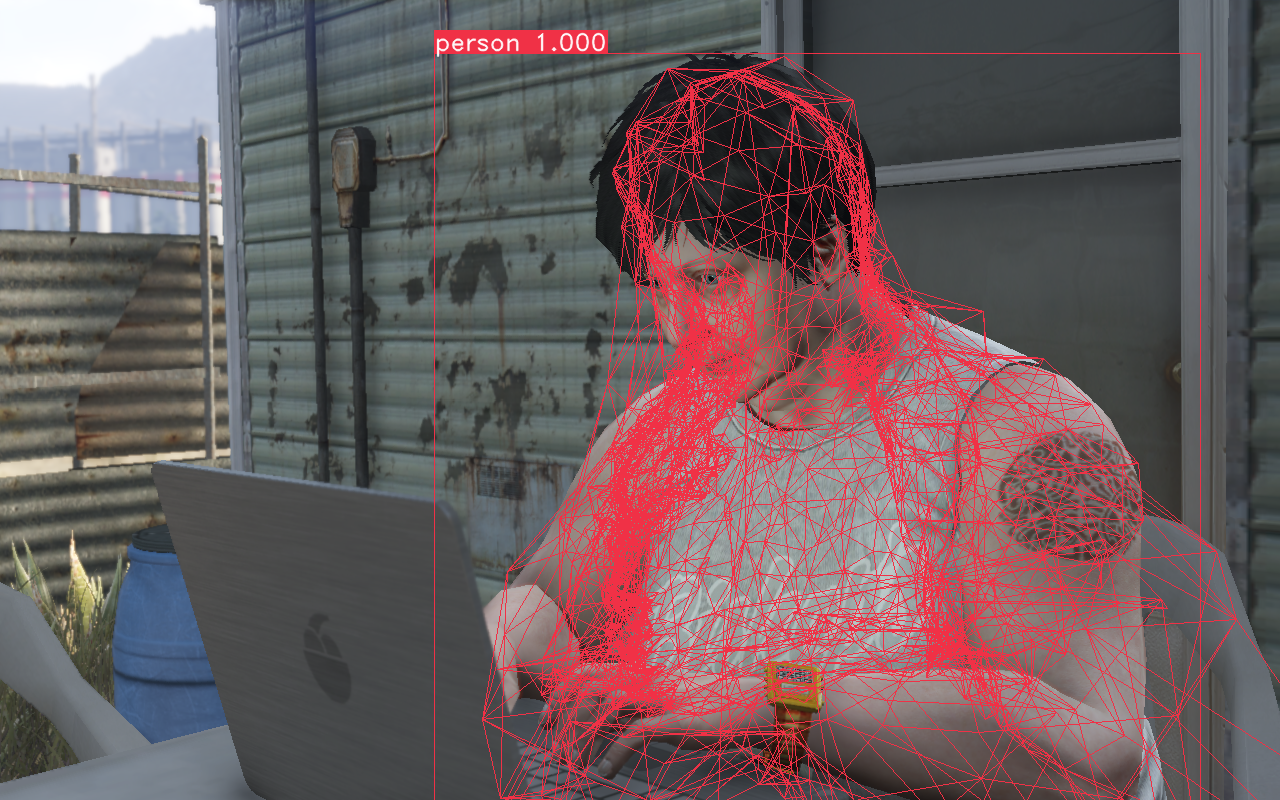} & 			\includegraphics[align=c, width=0.235\textwidth]{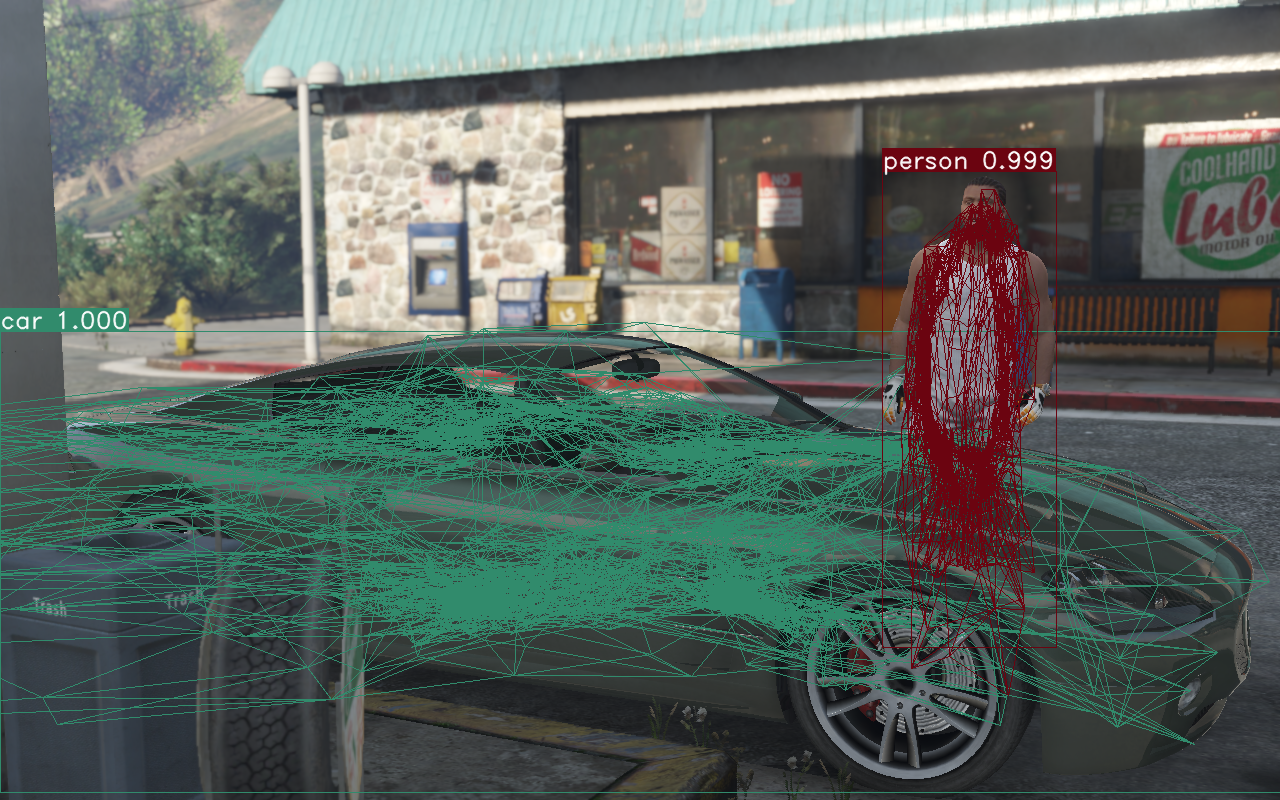} & 			\includegraphics[align=c, width=0.235\textwidth]{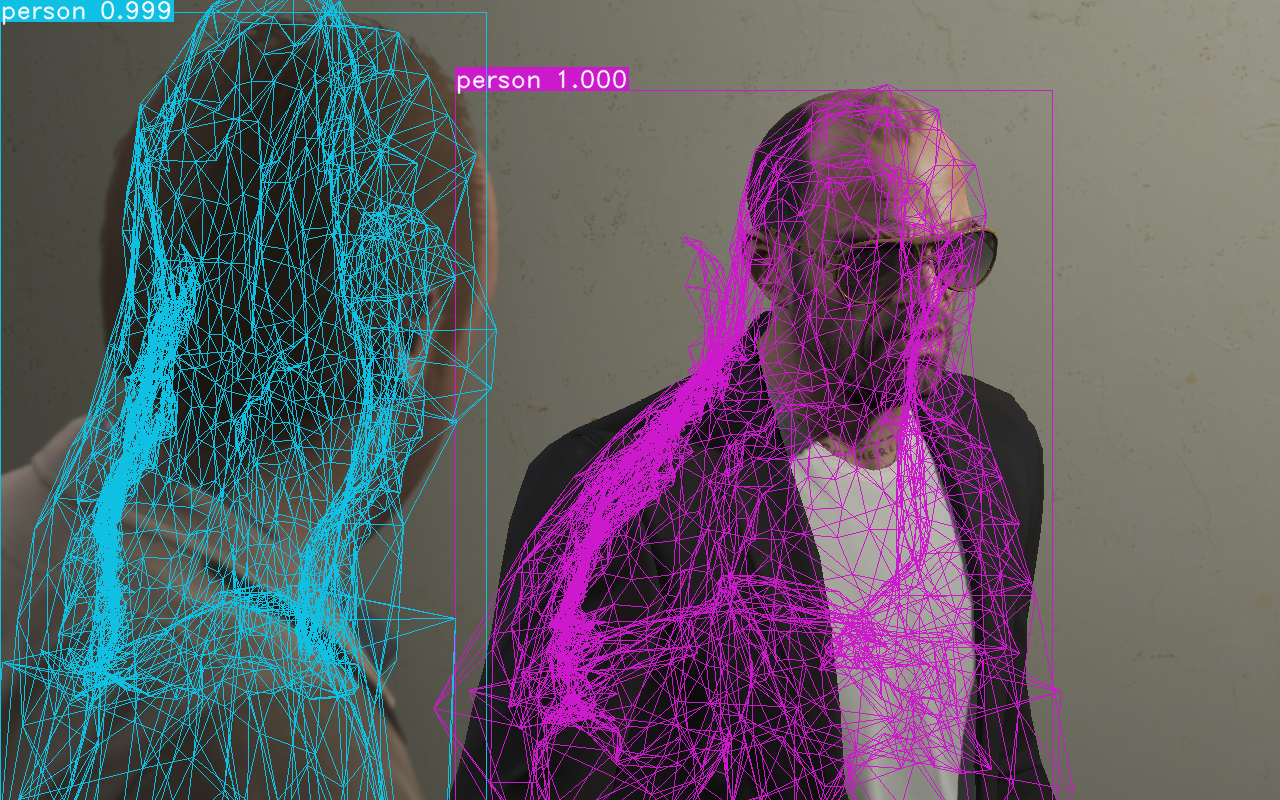} & 			\includegraphics[align=c, width=0.235\textwidth]{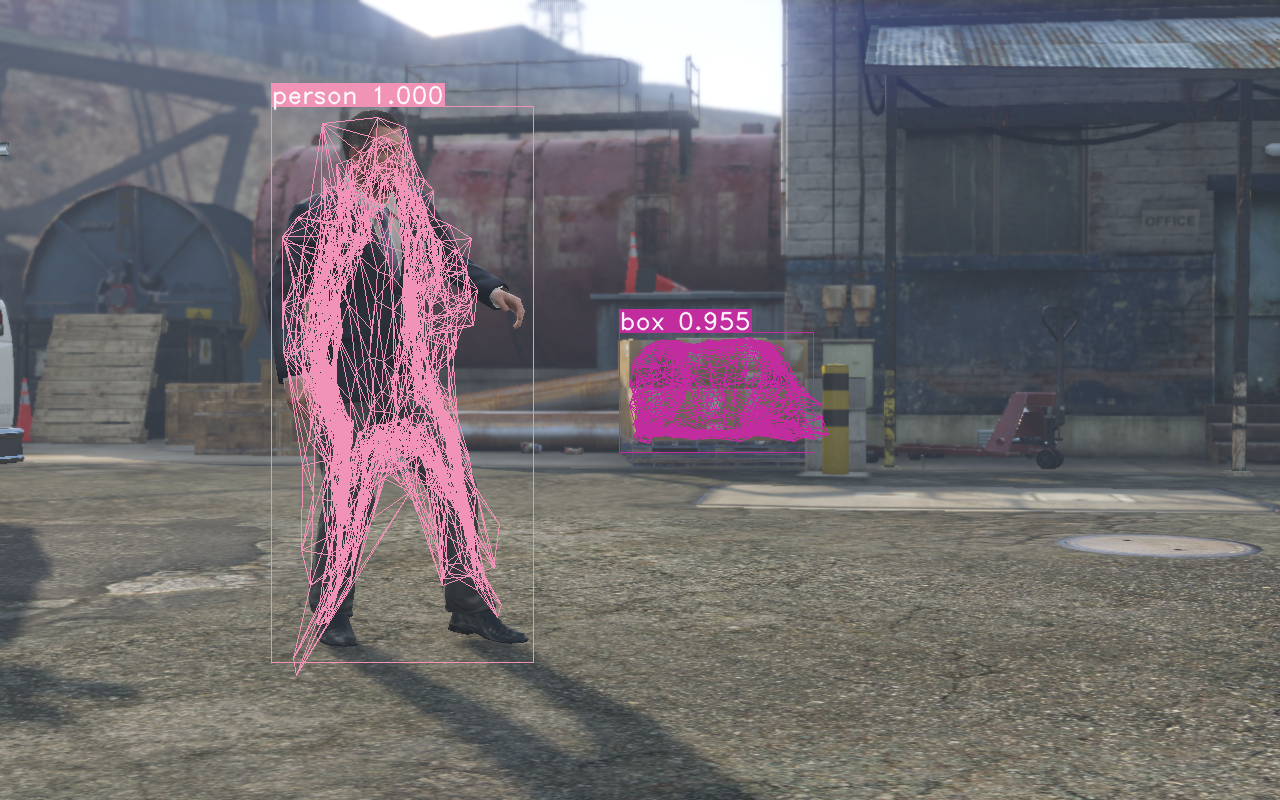}
        \end{tabular}
\end{minipage}
\vspace{-0.3cm}
\captionof{figure}{Visualization of 3D reconstructions on SAIL-VOS 3D.}
\label{fig:exp}
\vspace{-0.5cm}

\end{figure*}

Importantly, and different from prior work, instead of predicting meshes from a sphere~\cite{wang2018pixel2mesh}, our approach depends on an additional \textit{time dependent} reference mesh $\tilde{m}_{t-1}^i$.
This reference $\tilde{m}_{t-1}^i$ summarizes
information about an object $o_t^i$ obtained from the previous frame $I_{t-1}$. This enables a recurrent formulation suitable for mesh generation from videos by incorporating temporal information.

\subsection{Training}\label{sec:training}
We train our model using the dataset discussed in \secref{sec:data} which consists of sequences of image frames, object bounding box annotations, class labels, and corresponding meshes. In the following, we discuss the training details abstractly and leave specifics to Appendix~\secref{sec:supp_details}. %

To train the model parameters for mesh prediction, we use a gradient-based method to minimize the mesh loss based on differentiable mesh sampling~\cite{smith2019geometrics}. The loss includes the Chamfer and normal distance between two sets of sampled points each from the prediction and ground-truth mesh. Minimizing the mesh loss encourages the predicted meshes to be similar to the  ground-truth mesh. The training of bounding-boxes follows the standard Mask-RCNN procedure~\cite{he2017mask}. %

Different from standard end-to-end training, our training procedure consist of two stages:\\
\noindent\textbf{Stage 1:} %
We first pre-train the mesh refinement without taking temporal information into account. This is done by setting the reference mesh $\tilde{m}_{t-1}^i = \mT_{o_t^i}( \bar{M}_{c_t^i})$ $\forall \tilde{o}_{t-1}^i$.

\noindent\textbf{Stage 2:} We fine-tune our model to incorporate the temporal information of the reference mesh following~\equref{eq:reference}. Note that at training time, we do not unroll the recursion which would be computationally expensive.
Instead, we use an augmented ground-truth mesh of $o_t^i$ as an approximation. This is done by randomly rotating the mesh and adding Gaussian noise to the vertices. Without the augmentation, the ground-truth mesh will not resemble $\tilde{m}_{t-1}^i$ which leads to a mismatch at test time.

\section{Experimental Results}

In the following we first discuss the experimental setup, \ie, datasets, metrics and baselines. Afterwards we show quantitative and qualitative results on the proposed SAIL-VOS 3D dataset and the commonly used Pix3D dataset. %

\subsection{Experimental Setup}

We use the proposed SAIL-VOS 3D dataset to study instance-level video object reconstruction. Please see \secref{sec:datastat} for statistics of this data. We also evaluate on  Pix3D~\cite{sun2018pix3d} to assess the proposed method on real images.

\noindent\textbf{Evaluation metrics:} 
For evaluation we use   metrics commonly employed for  object detection. Specifically, we report the average precision for bounding box detection, \ie, 
$\text{AP}^{\text{box}}$, using an intersection over union  (IoU) threshold of 0.5. We also provide the average precision for mask prediction, \ie, 
$\text{AP}^{\text{mask}}$, using an IoU threshold of 0.5. 

Additionally, to evaluate 3D shape prediction we follow Gkioxari~\etal~\cite{gkioxari2019mesh} and compute the average area under the per-category precision-recall curve using  F1@0.3 with a threshold of 0.5, which we refer to via 
$\text{AP}^{\text{mesh}}$. %
Note, we consider a predicted mesh a true-positive if its predicted label is correct, not a duplicate detection, and its F1@0.3 $\geq 0.5$.
Here, F1@0.3 is the F-score computed using precision and recall. For this, precision is the percentage of points sampled from the predicted mesh that lie within a  distance of 0.3 to the ground-truth mesh. Conversely, recall is the percentage of points sampled from the ground-truth mesh that lie within a distance of 0.3 to the prediction. For SAIL-VOS 3D, we rescale the meshes such that the longest edge of the ground-truth mesh is 5. For Pix3D, following~\cite{gkioxari2019mesh}, we rescale the meshes such that the longest edge of the ground-truth mesh is 10.

For  more insights, we further evaluate the aforementioned metrics on subsets of the object instances. This includes a split based on objects' sizes (small -- area less than $32^2$, medium, and large -- area greater than $96^2$). %
We also split the objects based on  occlusion rate, \ie, an object is  heavily occluded if its occlusion rate is greater than $25\%$, otherwise it's slightly occluded. Finally, we split based on the length of the video clip, \ie, a clip is classified as long if the number of frames is more than 30, otherwise short.

\noindent\textbf{Baselines:} 
In addition to the developed method we study three baselines on the proposed SAIL-VOS 3D data:

\emph{MeshR-CNN}~\cite{gkioxari2019mesh} --  a single-view multi-object baseline. Its output representation is a mesh but the intermediate result is a voxel representation. %

\emph{Pixel2Mesh$\dagger$}  --  The original Pixel2Mesh~\cite{wang2018pixel2mesh} is a single-view single-object baseline. To deal with multiple objects we use Mask-RCNN and attach  Pixel2Mesh as an ROI head. We refer to this adjustment via Pixel2Mesh$\dagger$. We use the implementation by Gkioxari~\etal~\cite{gkioxari2019mesh} for this extension.
We also considered as a baseline Pixel2Mesh++~\cite{wen2019pixel2mesh++}, a multi-view single-object method. The fact that it is a single-object method makes a simple extension challenging as we would need to incorporate a tracking procedure. Moreover, Pixel2Mesh++ assumes camera matrices to be known which is not always feasible.

\emph{Pix2Vox$\dagger$} -- The original Pix2Vox~\cite{xie2019pix2vox} is a  multi-view single-object baseline which uses a voxel representation. Similar to Pixel2Mesh$\dagger$, to enable multiple object reconstruction, we extend  the original Pix2Vox by attaching it to a Mask-RCNN. %
Similar to Pixel2Mesh++ it is non-trivial to extend Pix2Vox to the multi-object setting. It would require to  include object tracking. %

\begin{figure}[t]
	\centering
	\begin{minipage}[t]{\linewidth}
    \begin{minipage}[t]{0.48\linewidth}
    	\centering Input \\
    	\includegraphics[width=1.65in,height=1.in]{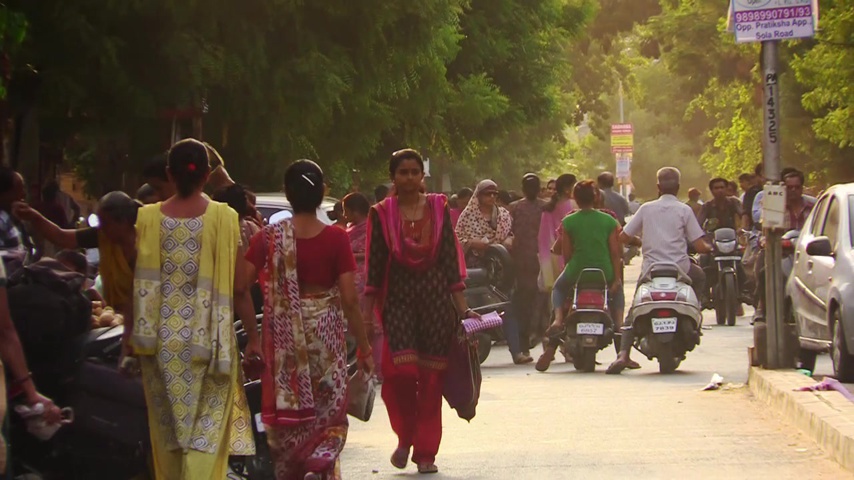}\\
    		\vspace{0.1cm} 
    	\includegraphics[width=1.65in,height=1.in]{Figs/davis/00001.jpg} \\
	\end{minipage}
\hfill
	 \begin{minipage}[t]{0.48\linewidth}
		\centering Predicted Shape \\
		\includegraphics[width=1.65in,height=1.in]{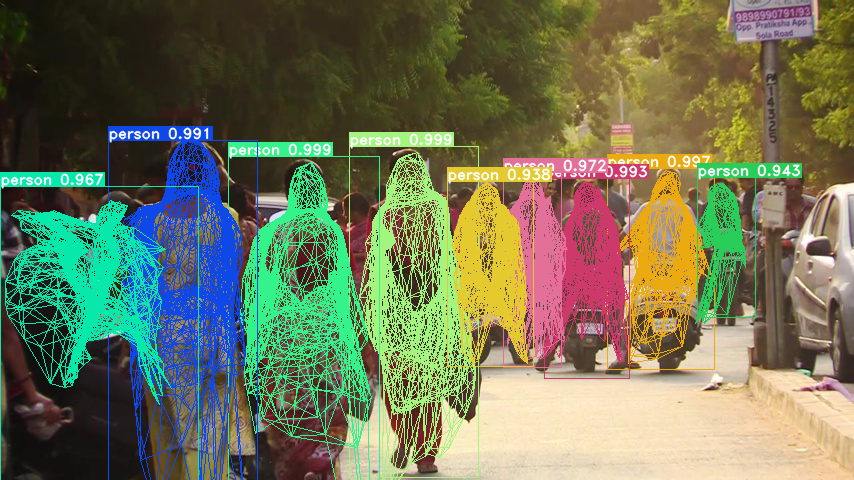}\\
			\vspace{0.1cm}
		\includegraphics[width=1.65in,height=1.in]{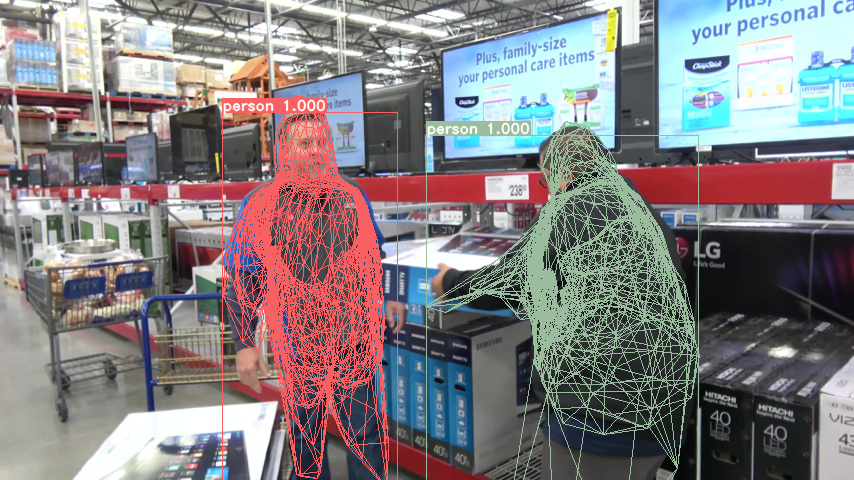}\\
	\end{minipage}
	\vspace{-0.1in}
\end{minipage}
	\caption{Predicted shapes of the SAIL-VOS 3D trained Video2Mesh model tested on DAVIS.}\label{fig:davis}
	\vspace{-0.15in}
\end{figure}

\begin{figure}[t]
	\includegraphics[align=c, width=\linewidth,height=1.in]{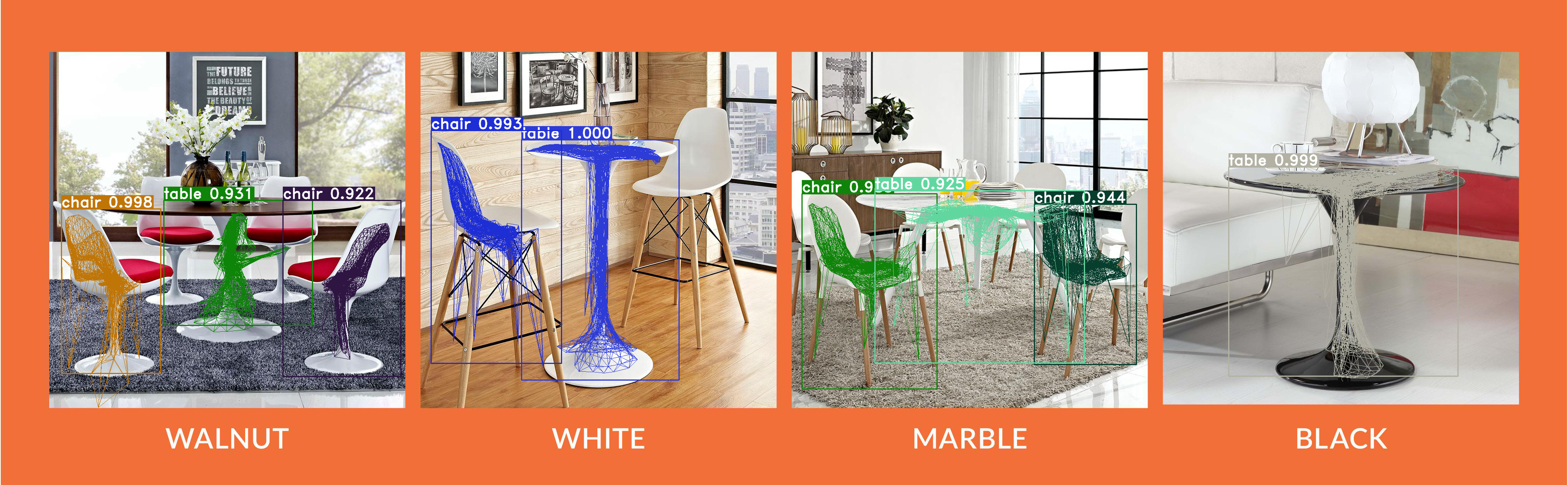}\vspace{0.05cm}\\
	\includegraphics[align=c, width=0.235\textwidth,height=1.in]{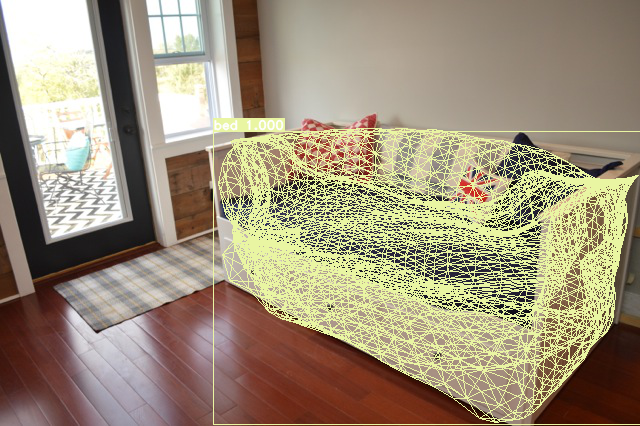} 	\includegraphics[align=c, width=0.235\textwidth,height=1.in]{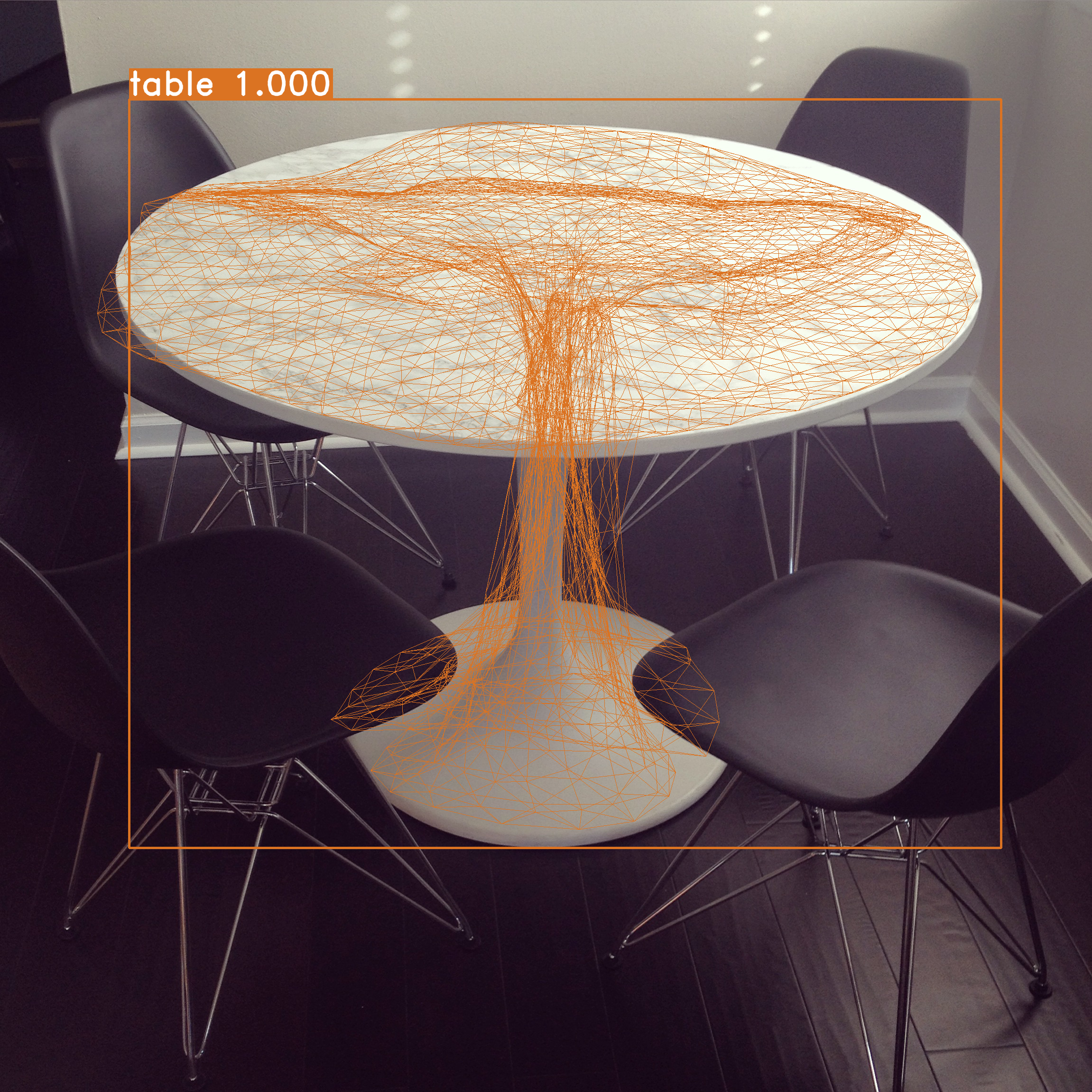} \\
	\vspace{-0.3cm}
	\captionof{figure}{Visualization of 3D reconstructions from our Video2Mesh method on Pix3D.}
	\label{fig:shapenet}
	\vspace{-0.6cm}
	
\end{figure}

\subsection{SAIL-VOS 3D Dataset}

SAIL-VOS 3D consists of 484 videos made up of 215,703 image frames. The data is split into 162 training videos (72,567 images, 754,895 instances) and 41 clips for validation (1,305 images, 12,199 instances). We leave the rest in test-dev/test-challenge sets for future use. 
Following Hu \etal~\cite{HuCVPR2019} we use the 24 classes with an occlusion rate less than 0.75  for experimentation. To compute the mean shape, we average  the occupancy grids in the object coordinate system over all the instances of a class using the training set. We binarize the averaged occupancy values, find the largest connected component, perform marching cubes to get the mesh, and simplify the mesh such that it contains about 4000 faces using~\cite{huang2018robust}. %

{\noindent\bf Quantitative results:} Quantitative results for the SAIL-VOS 3D dataset are provided in \tabref{tab:exp}. Note that we first train and fix the box and mask heads and then train different baselines for predicting meshes. On SAIL-VOS 3D we find that methods which predict shape via a mesh representation (Pixel2Mesh$\dagger$, MeshR-CNN, and our Video2Mesh) perform better than  methods which predict using a voxel representation (Pix2Vox$\dagger$). We observe that our method outperforms the best baseline MeshR-CNN~\cite{gkioxari2019mesh} in terms of $\text{AP}^{\text{mesh}}$.  More notably, our method also outperforms baselines in slightly and heavily occluded subsets as well as long clips. This is to be expected, as our approach utilizes temporal information which helps to reason about occluded regions. To confirm this, we conduct an ablation study.

{\noindent\bf Ablation study:} We study the importance of the mean shape (abbreviated via `MS'), whether to estimate transformation between mean shape and current frame (abbreviated via `T') and the temporal prediction (abbreviated via `Temp.'). 
The results are summarized in \tabref{tab:ablation}. Note that we fix the box and mask head for this ablation study, which ensures a proper evaluation of our mesh head modifications. We observe use of the mean shape to lead to slight improvements (Row 2 \vs Row 1). Having transformation enabled improves results more significantly (Row 3 \vs Row 2), while the temporal prediction leads to further increase in performance (Row 4 \vs Row 3).

{\noindent\bf Qualitative results on SAIL-VOS 3D:} Qualitative results are shown in \figref{fig:exp}. We compare to the runner-up approach MeshR-CNN. Due to the use of a mean shape, we observe  the 3D reconstructions of our method to be  smoother.

{\noindent\bf Qualitative results on Real Data:} We tested a SAIL-VOS 3D %
trained model (trained on person only) on a real video from the DAVIS dataset~\cite{Pont-Tuset_arXiv_2017} and show a qualitative example in \figref{fig:davis}. We find that predicted shapes are reasonable even though the model is trained on synthetic data.

\section{Transferring to Pix3D}\label{sec:transffering}

To assess the use of  synthetic data we study transferability. Specifically, we use the SAIL-VOS 3D pretrained model and finetune on Pix3D~\cite{sun2018pix3d} using different amounts of Pix3D training data. Pix3D~\cite{sun2018pix3d} is an image dataset. Consequently, we cannot study  temporal aspects of our  model. We use as conditional input the mean shape rather than the prediction from the previous frame. 
For evaluation we follow Gkioxari~\etal~\cite{gkioxari2019mesh} and report the AP metrics in \tabref{tab:transferring}.  More details regarding experimental setup and implementation are given in Appendix~\ref{sec:supp_details}.  Qualitative results of our Video2Mesh method trained on all Pix3D data are provided in \figref{fig:shapenet}. 

\noindent{}In the low-data regime where only 10\%, 20\%, and 50\% real mesh data (Pix3D) are available, pretraining on synthetic data is beneficial: the model pretrained on SAIL-VOS 3D outperfoms the model without pretraining. Note that depth data may also be a useful cue to facilitate transferability which we didn't study yet.

\begin{table}[t]
	\centering
	\caption{Results of transferring SAIL-VOS 3D to Pix3D ($\text{AP}^{\text{mesh}}$).}
	\label{tab:transferring}
	\vspace{-0.4cm}
	\setlength{\tabcolsep}{3pt}
	\begin{tabular}{@{} lcccc@{} }
		\hline
		\% of Pix3D data & 10\% & 20\% & 50\% & 100\% \\
		\hline 
		Ours & 9.0 & 16.8 & 36.6 & \bf 43.7\\
		Ours pretrained w/ SAIL-VOS 3D & \bf 16.4 & \bf 22.5 & \bf 38.0 & 42.8\\
		\hline
	\end{tabular}
			\vspace{-0.5cm}
\end{table}
\vspace{-0.2cm}
\section{Conclusion}
\vspace{-0.2cm}
We introduce  SAIL-VOS 3D  and develop a  baseline for 3D mesh reconstruction from video data. In a first study we observe temporal data to aid reconstruction. We hope SAIL-VOS 3D facilitates further research in this direction. %

\noindent\textbf{Acknowledgements:} This work is supported in part by NSF under Grant \#1718221, 2008387, 2045586, MRI \#1725729, and NIFA award 2020-67021-32799, UIUC, Samsung, Amazon, 3M, and Cisco Systems Inc.\ (Gift Award CG 1377144 - thanks for access to  Arcetri). 

\clearpage
{\small
\bibliographystyle{ieee_fullname}
\bibliography{egbib}
}
\clearpage
\appendix
\renewcommand{\thetable}{A\arabic{table}}
\setcounter{table}{0}
\setcounter{figure}{0}
\renewcommand{\thetable}{A\arabic{table}}
\renewcommand\thefigure{A\arabic{figure}}
\renewcommand{\theHtable}{A.Tab.\arabic{table}}%
\renewcommand{\theHfigure}{A.Abb.\arabic{figure}}%
\renewcommand\theequation{A\arabic{equation}}
\renewcommand{\theHequation}{A.Abb.\arabic{equation}}%

\newcommand*{\dictchar}[1]{
    \clearpage
    \twocolumn[
    \centerline{\parbox[c][3cm][c]{\textwidth}{%
            \centering
            \fontsize{14}{14}
            \selectfont
            {#1}}}]
}

\onecolumn
{\centering \Large \textbf{Appendix: SAIL-VOS 3D: A Synthetic Dataset and Baselines for Object Detection and 3DMesh Reconstruction from Video Data}}

\section{Additional Details}\label{sec:supp_details}

\noindent\textbf{Mesh loss:} 
Recall, in~\secref{sec:training} %
we optimize the model parameters to minimize a mesh loss which encourages the predicted meshes to be similar to the ground-truth. In this section, we  describe the details on how to compute this loss. 
Following Mesh R-CNN~\cite{gkioxari2019mesh}, the mesh loss is the sum of three distances, the Chamfer distance, the normal distance, and an edge regularizer, each of which we discuss in the following.

Given a predicted mesh $m_t^i$ and its corresponding ground-truth mesh $m_t^{i,*}$, differentiable sampling~\cite{smith2019geometrics} is used to extract point clouds with normal vectors, $P$ and $P^*$, from each of the meshes. 
We sample 5,000 3D points from the meshes to compute the distances,~\ie, $|P| = |P^*| =  5000$ and for each point $p\in P$, $p^\ast \in P^\ast$ we have $p\in\mathbb{R}^3$ and $p^\ast\in\mathbb{R}^3$. 
We use those sets to compute the three losses as follows.

{\noindent \it Chamfer distance:} 
\bea
L_{\text{cham}}(P, P^*) = \frac{1}{|P|} \sum_{p \in P} \min_{p^* \in P^*} \norm{p-p^*}^2
+ \frac{1}{|P^*|} \sum_{p^* \in P^*} \min_{p\in P}\norm{p^*-p}^2.
\eea
{\noindent \it Normal distance:}
\bea
L_{\text{norm}}(P,P^*) = \frac{1}{|P|} \sum_{p \in P} \min_{p^* \in P^*} |\mathbf{u}_p^{\intercal} \mathbf{u}_{p^*}|
+ \frac{1}{|P^*|} \sum_{p^* \in P^*} \min_{p\in P} |\mathbf{u}_{p^*}^{\intercal} \mathbf{u}_p|,
\eea
where $\mathbf{u}_p$ corresponds to the unit normal vector of a point $p\in P$ on a mesh surface. 

{\noindent \it Edge regularizer:} 
\be
L_{\text{edge}}(V,E) = \frac{1}{|E|} \sum_{(v,v') \in E} \norm{v-v'}^2,
\ee
where 
$V$ and $E \subseteq V \times V$ denote the set of vertices and edges of the predicted mesh. Here, $v\in V$ is a 3D vertex of the mesh. 

The overall mesh loss for our model is a weighted sum of the Chamfer distance, normal distance, and edge regularizer on the predicted mesh at each stage of the mesh-refinement.

{\noindent \bf Mean mesh alignment:}
Recall, in \equref{eq:reference} we introduce a transformation matrix $T_{o_t^i}$ to align the class-specific mean-mesh to the current object. More formally, we parameterize $T_{o_t^i}$ using a deep-net, which uses as input the ROI features $R_{o_t^i}$ and regresses to a rotation matrix with three degrees of freedom. We do not consider translation here, as the mesh is centered in a bounding box. To learn this transformation matrix, we train the parameters of $T_{o_t^i}$ to minimize the Chamfer distance and the normal distance between the mean-mesh and the ground-truth mesh. Note that we exclude the edge regularizer as a rotation maintains the edges' length. 

{\noindent \bf Bounding box depth extent prediction:}
We denote the depth extent via $dz$ and refer to the depth via $z_c$. We obtain both from the ground-truth mesh via
\be
dz = z_\text{far}-z_\text{near}, \quad
z_c=\frac{(z_\text{far}+z_\text{near})}{2},
\ee
where $z_\text{far} = \max_{v\in V} v_z$ and $z_\text{near} = \min_{v\in V}v_z$. Here, $v_z$ is the z-coordinate of vertex $v\in V$.
Instead of predicting $dz$, following MeshR-CNN~\cite{gkioxari2019mesh}  we predict the scale normalized depth extent 
\be
\bar{dz}=\frac{dz}{z_c}\frac{f}{h},
\ee
where $f$ is the focal length and $h$ is the height of the bounding box. Following Mesh R-CNN~\cite{gkioxari2019mesh}, we assume $z_c$ to be given during inference.

{\noindent \bf Implementation details:}
The model is optimized using SGD with a learning rate of 0.02  and momentum of 0.9. For SAIL-VOS 3D, we train each stage using 10,000 iterations . We use a batch size of 32 on 8 Tesla V100 GPUs. We use a  ResNet-50 model pretrained on COCO for initialization of the backbone, RPN, the bounding box head and the mask head. %
For all the experiments on Pix3D, we use the COCO pretrained model as the initialization of the parameters in the backbone, RPN, the box head and the mask head except for the class-specific top layers. The parameters  of the class-specific top layers are randomly initialized. For the experiments in ~\secref{sec:transffering} where we use the SAIL-VOS 3D pretrained model, we initialize the parameters in the mesh head from the SAIL-VOS 3D pretrained model.

\section{Statistics of Clip Length on SAIL-VOS 3D}

\begin{figure*}[h]
	\vspace{-0.3cm}
	\centering
	\includegraphics[width=0.8\textwidth]{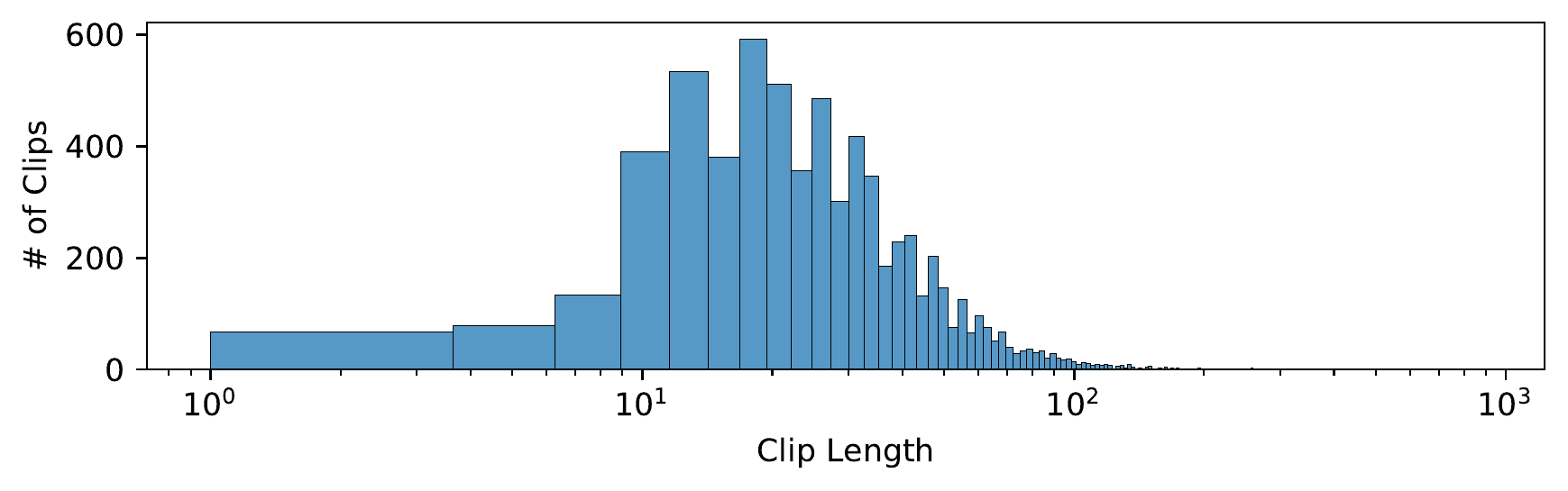}
	\vspace{-0.3cm}
	\caption{Clip length distribution of SAIL-VOS 3D dataset.
	}
	\label{fig:clip}
	\vspace{-0.3cm}
\end{figure*}

In \figref{fig:clip}, we plot the distribution of clip length of the SAIL-VOS 3D datasets. There are 6807 clips in total. There are around 14 clips per video. The average clip length is 34.6. Maximum clip length is 876 and minimum length is 1.
  
\section{More Qualitative Results}

\begin{figure*}[h]
\vspace{0.5cm}
	\begin{minipage}[h]{\textwidth}
	\setlength{\tabcolsep}{1pt}
    \renewcommand{\arraystretch}{0.95}
        \begin{tabular}{ccccc}
        Input &
        \includegraphics[align=c, width=0.235\textwidth]{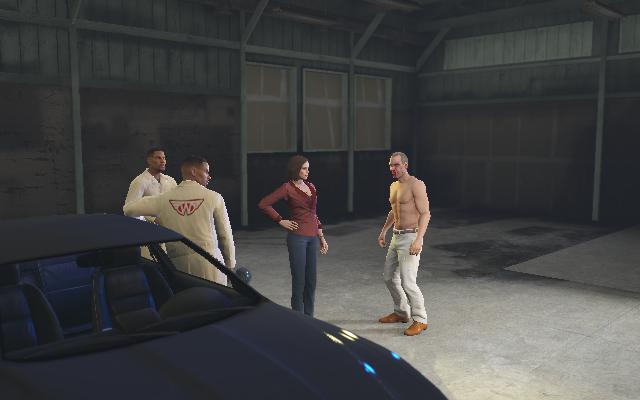} & 			\includegraphics[align=c, width=0.235\textwidth]{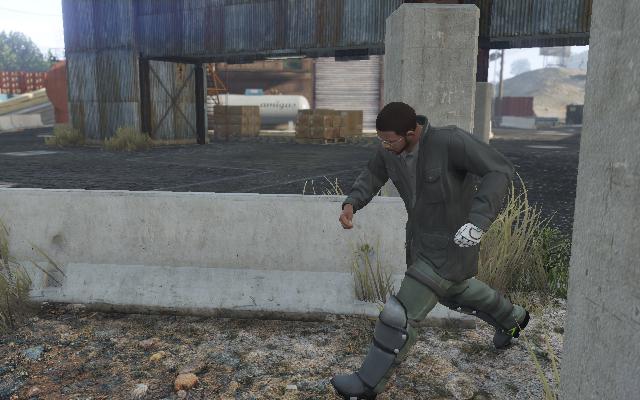} & 			\includegraphics[align=c, width=0.235\textwidth]{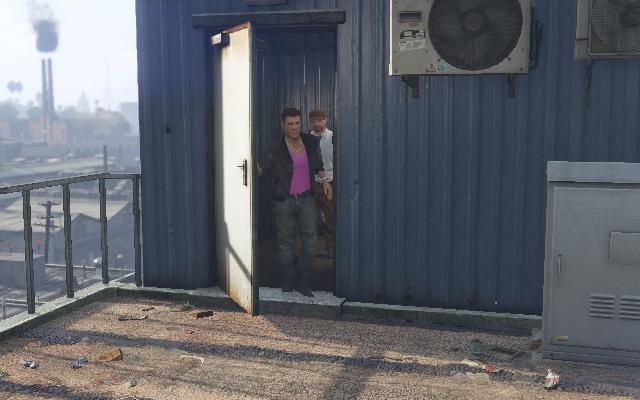} & 			\includegraphics[align=c, width=0.235\textwidth]{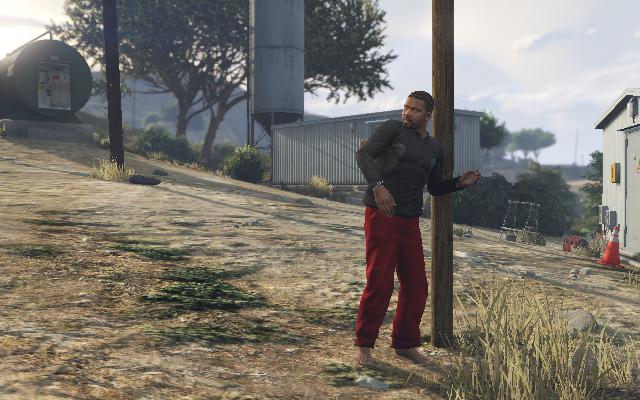} \\
        Results & 			\includegraphics[align=c, width=0.235\textwidth]{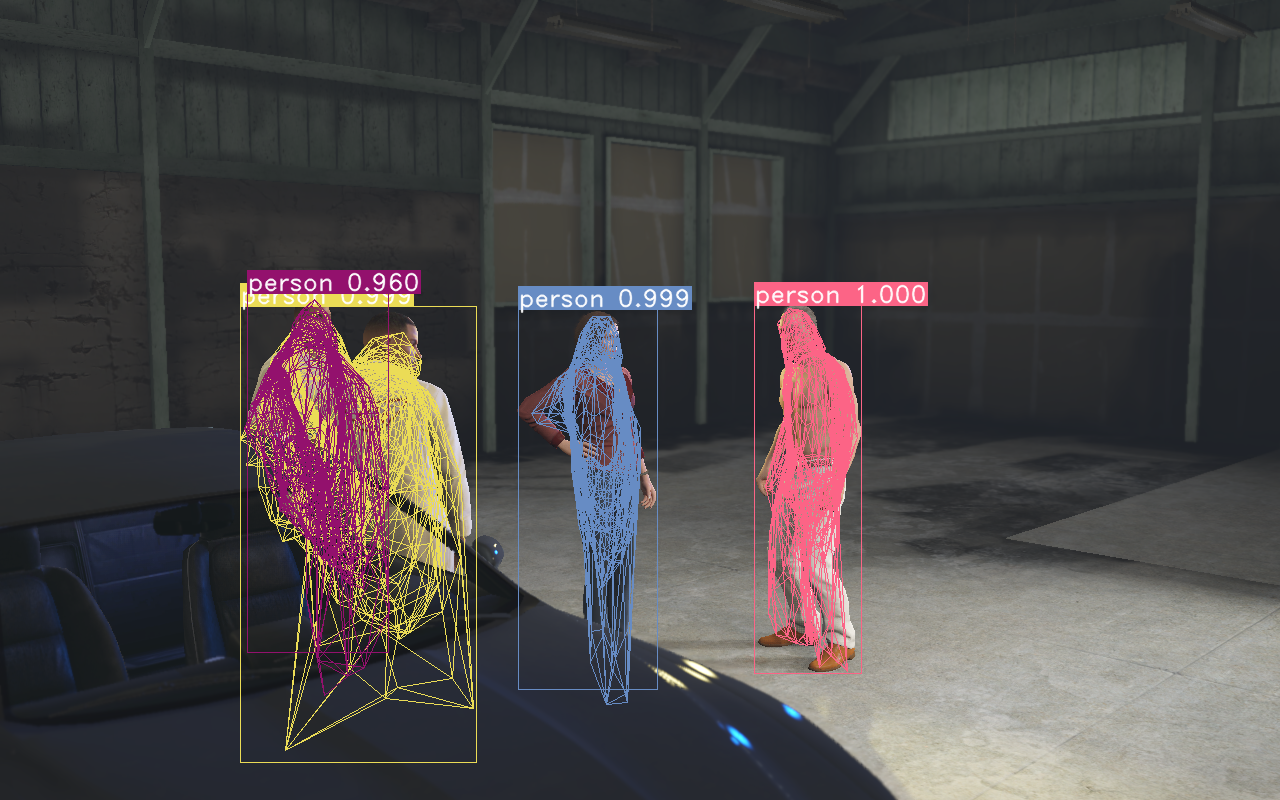} & 			\includegraphics[align=c, width=0.235\textwidth]{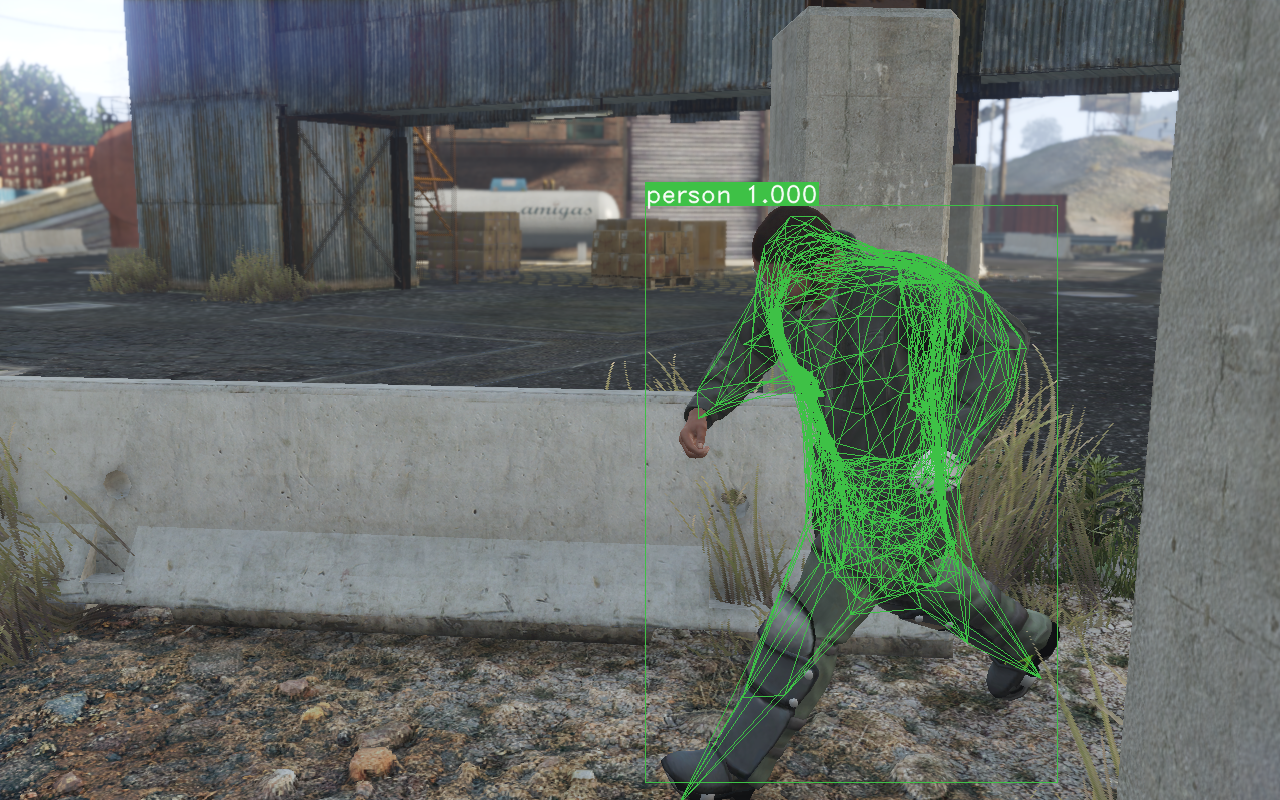} & 			\includegraphics[align=c, width=0.235\textwidth]{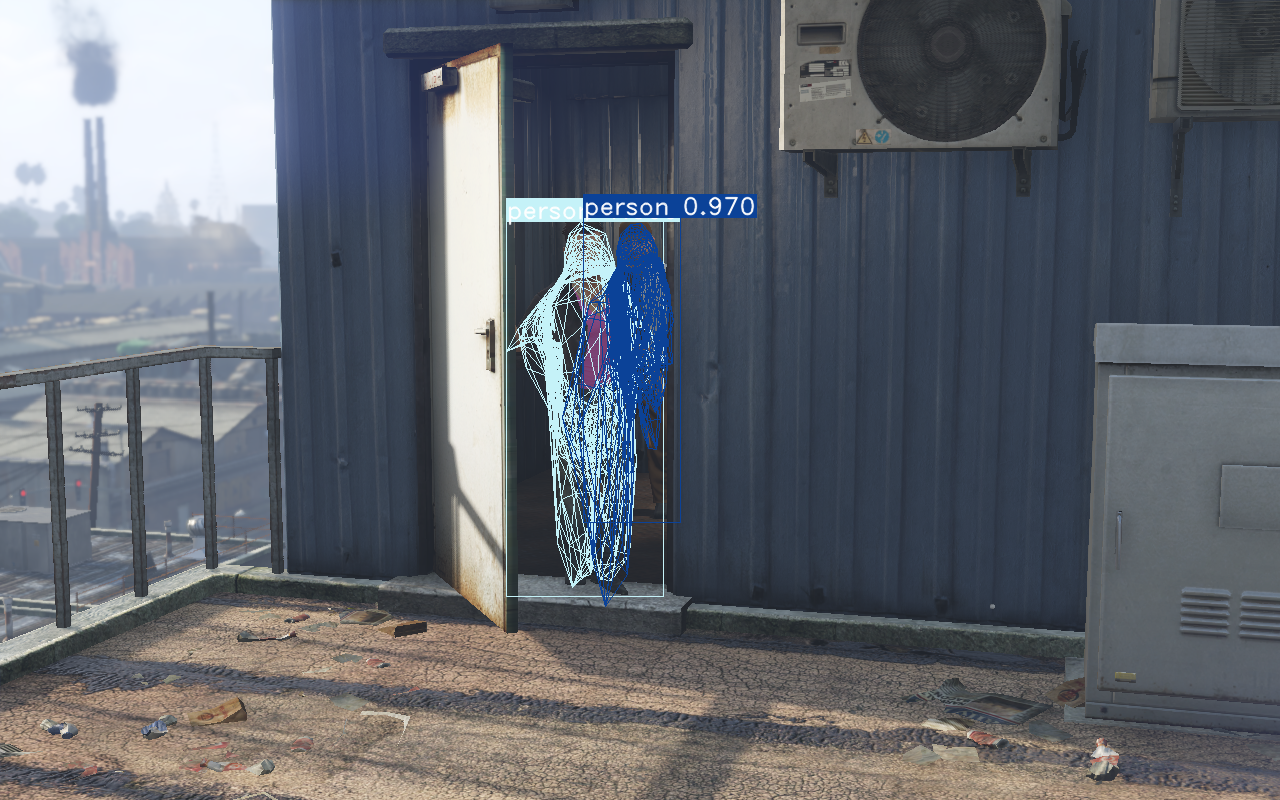} & 			\includegraphics[align=c, width=0.235\textwidth]{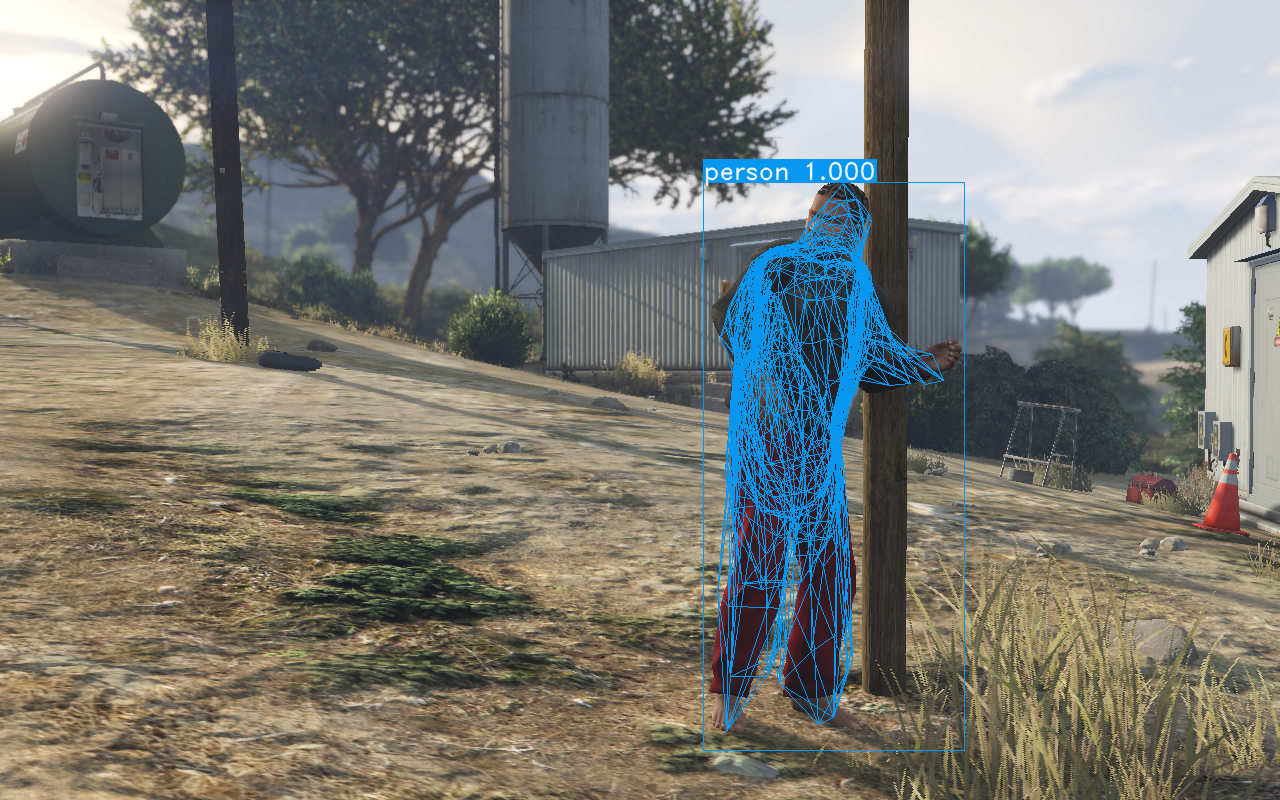}
        \end{tabular}
\end{minipage}
\vspace{-0.3cm}
\captionof{figure}{More qualitative results from Video2Mesh on SAIL-VOS 3D.}
\label{fig:supp_qual1}

\end{figure*}

\begin{figure*}[h]
\vspace{0.5cm}
	\begin{minipage}[h]{\textwidth}
	\setlength{\tabcolsep}{1pt}
    \renewcommand{\arraystretch}{0.95}
        \begin{tabular}{ccccc}
        Input &
        \includegraphics[align=c, width=0.235\textwidth]{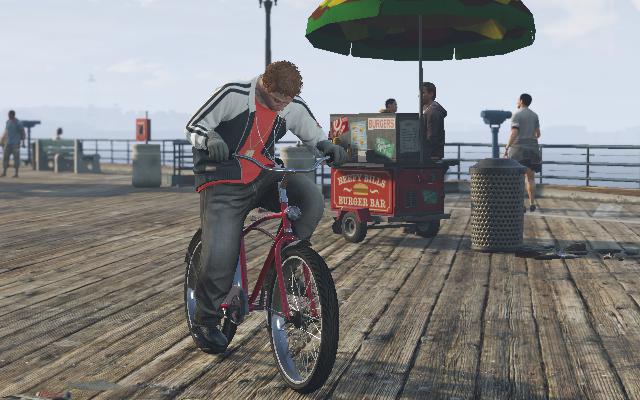} & 			\includegraphics[align=c, width=0.235\textwidth]{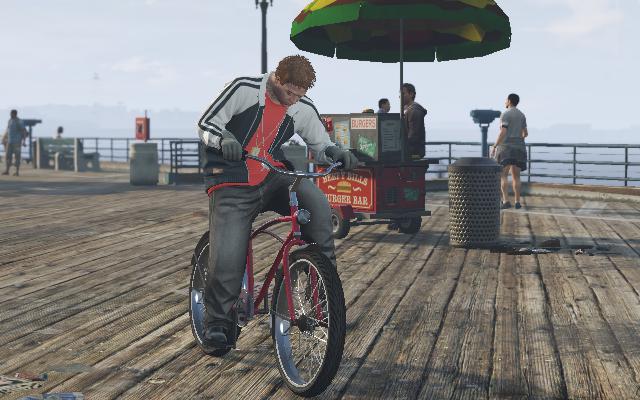} & 			\includegraphics[align=c, width=0.235\textwidth]{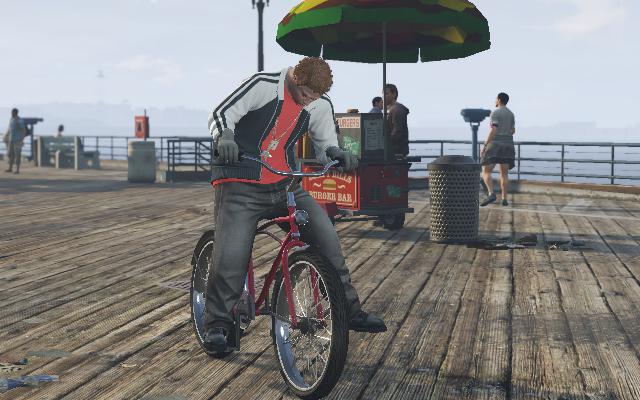} & 			\includegraphics[align=c, width=0.235\textwidth]{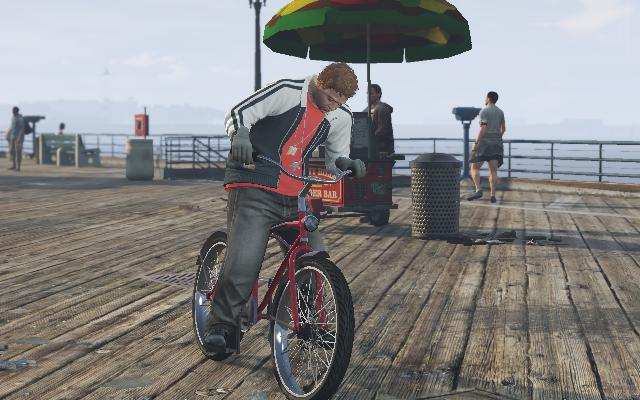} \\
        Results & 			\includegraphics[align=c, width=0.235\textwidth]{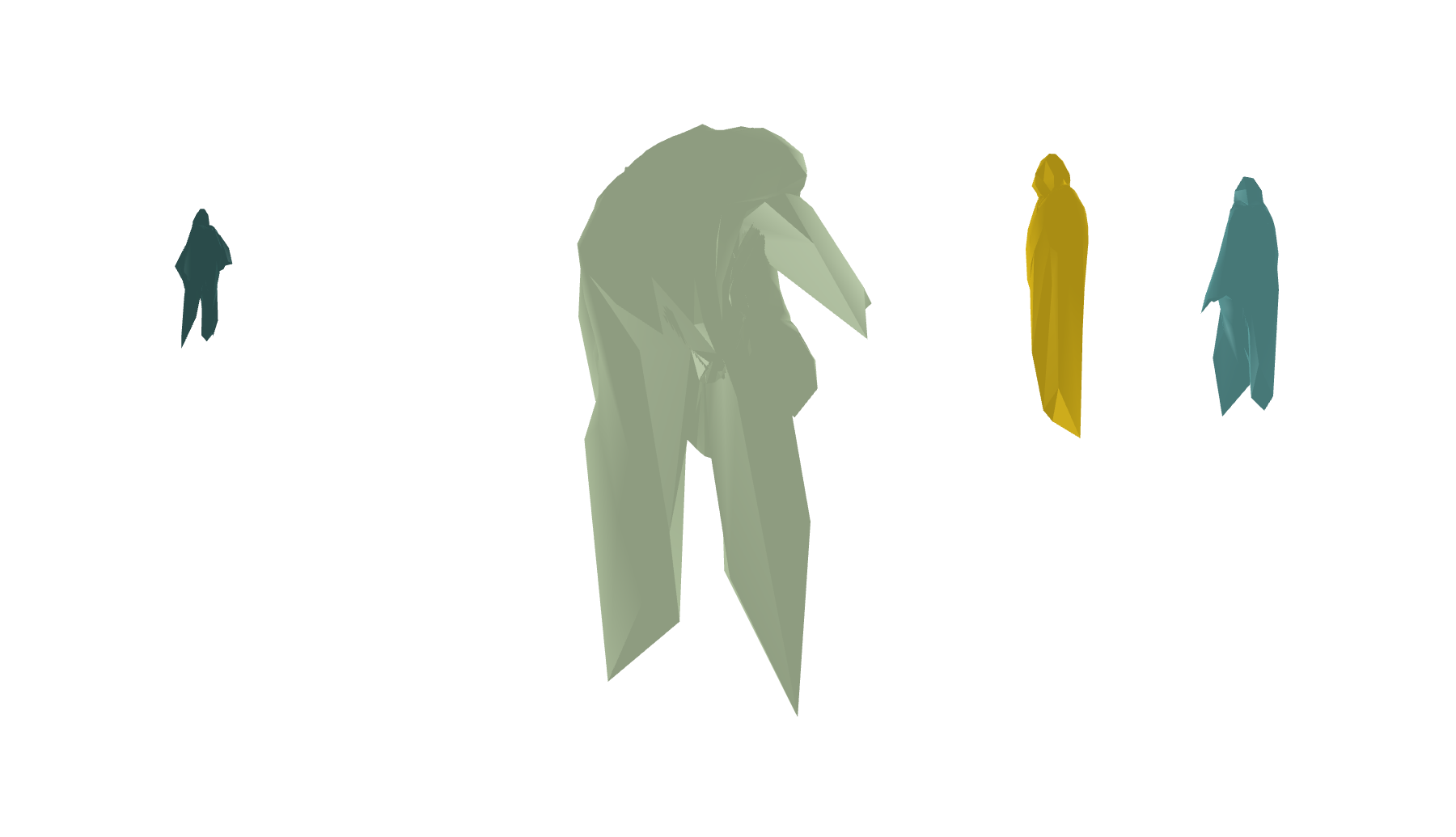} & 			\includegraphics[align=c, width=0.235\textwidth]{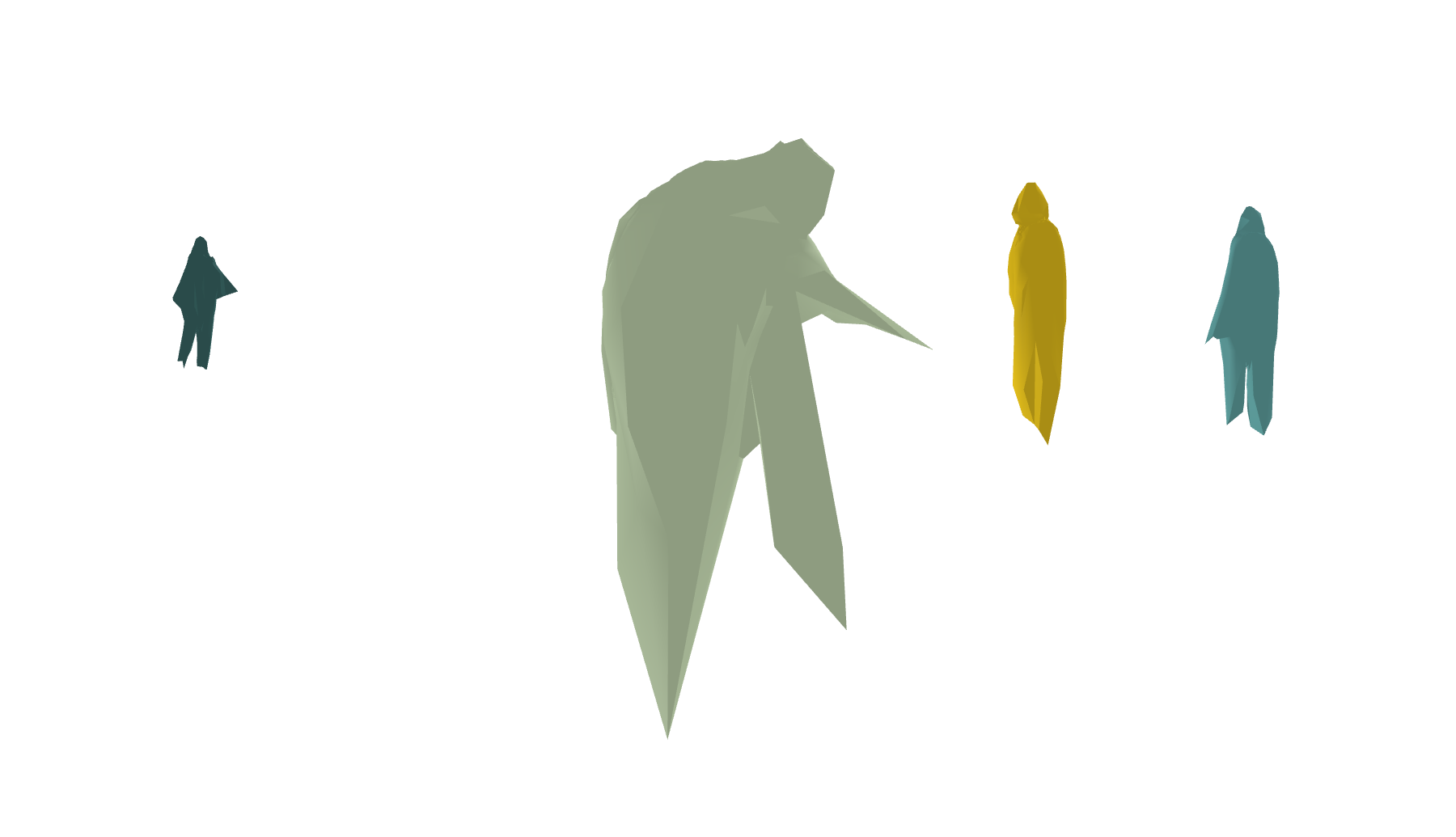} & 			\includegraphics[align=c, width=0.235\textwidth]{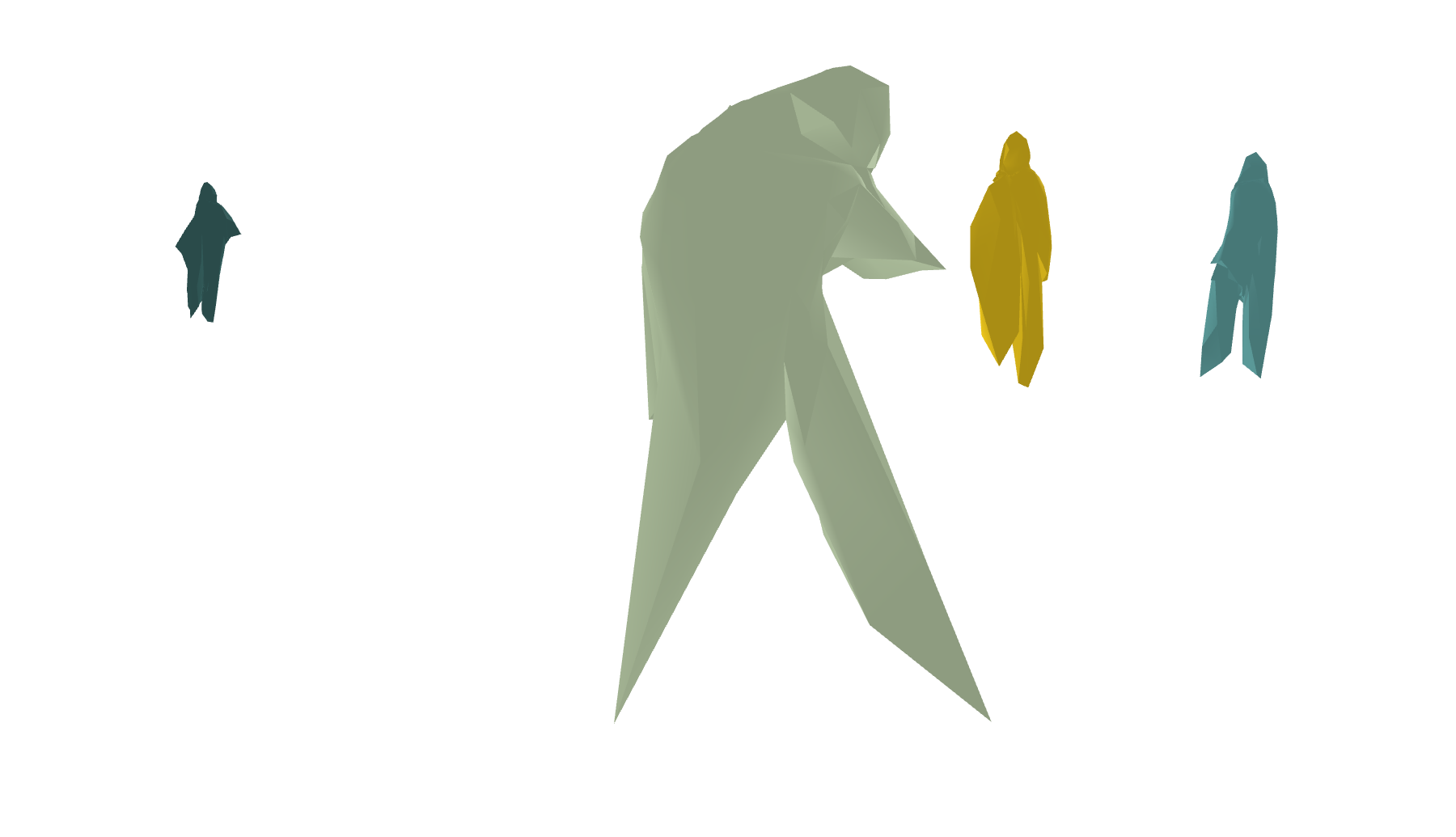} & 			\includegraphics[align=c, width=0.235\textwidth]{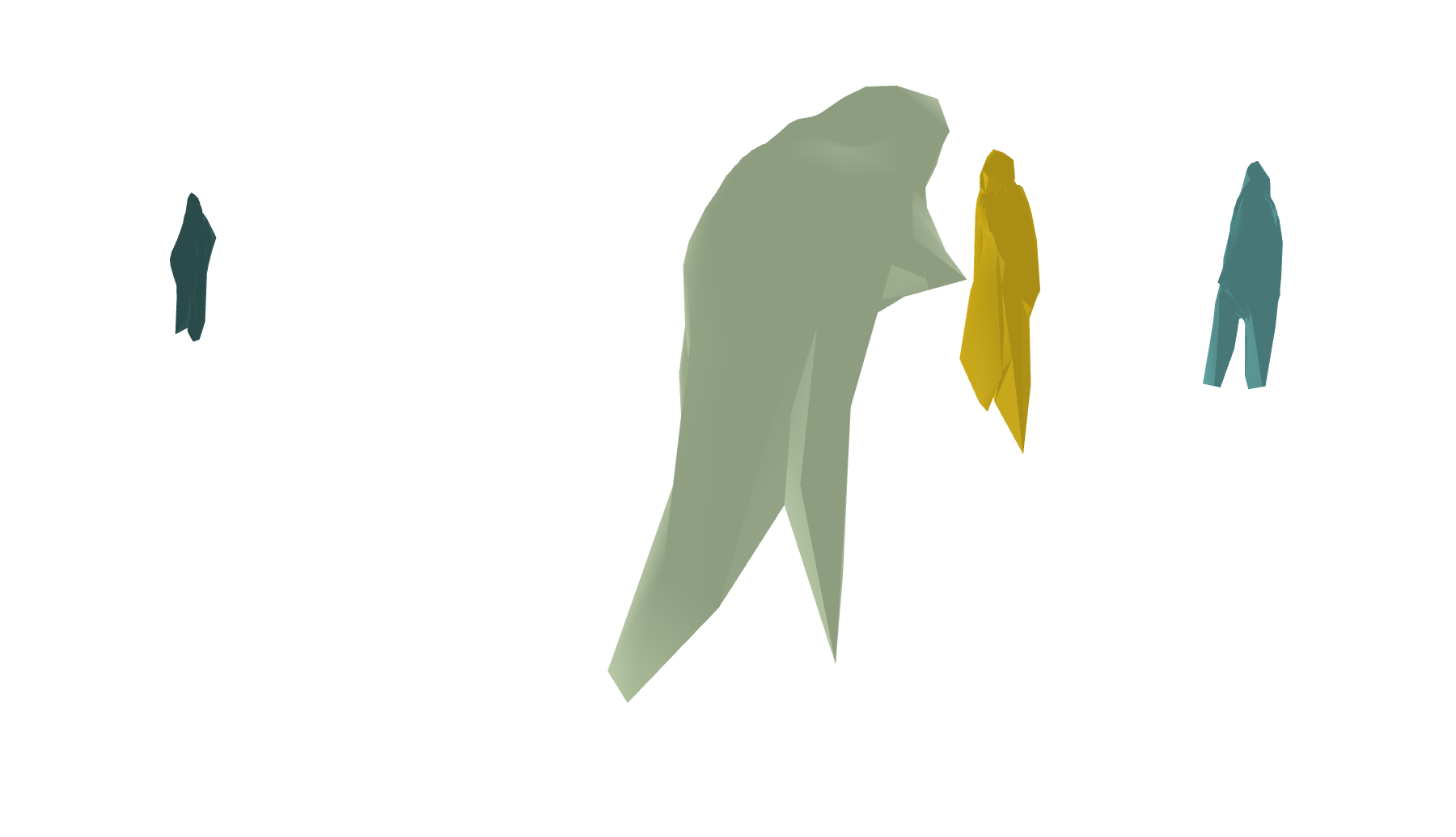}
        \end{tabular}
\end{minipage}
\captionof{figure}{More qualitative results from Video2Mesh on SAIL-VOS 3D. We show a video sequence here.}
\label{fig:supp_video}
\vspace{-0.5cm}

\end{figure*}

We present more qualitative results  of the proposed Video2Mesh method on SAIL-VOS 3D in~\figref{fig:supp_qual1}. The video results are visualized in~\figref{fig:supp_video}. We use the same colors to highlight the tracked objects across frames. Our method can detect and track multiple instances in a video sequence as shown in \figref{fig:supp_video} and reconstruct the 3D shapes as shown in \figref{fig:supp_qual1} and \ref{fig:supp_video}.

\end{document}